\pdfoutput=1

\documentclass[11pt]{article}

\usepackage[]{ACL2023}

\usepackage{times}
\usepackage{latexsym}
\usepackage{graphicx}
\usepackage{subfig}
\usepackage{booktabs}
\usepackage[T1]{fontenc}

\usepackage[utf8]{inputenc}
\usepackage{arabtex}
\usepackage{amsmath}
\usepackage{utf8}
\setcode{utf8}

\usepackage[]{mdframed}

\usepackage{microtype}

\usepackage{inconsolata}
\usepackage{textcomp}
\usepackage{multirow}

\newcommand{\proj}{mKGC}
\newcommand{\system}{\texttt{mRAKL}}

 \title{\texttt{mRAKL}: \underline{M}ultilingual \underline{R}etrieval-\underline{A}ugmented \underline{K}nowledge Graph Construction for \underline{L}ow-Resourced Languages.}

\author{Hellina Hailu Nigatu \\ UC Berkeley \\ hellina\_nigatu@berkeley.edu
        \And  Min Li  \\ Apple \\ min\_li6@apple.com
        \And Maartje ter Hoeve \\ Apple \\ m\_terhoeve@apple.com
        \AND Saloni Potdar \\ Apple \\ s\_potdar@apple.com
        \And Sarah E. Chasins \\UC Berkeley \\ schasins@berkeley.edu
}

\begin{document}
\maketitle
\begin{abstract}
\end{list}
    Knowledge Graphs represent real-world entities and the relationships between them. 
Multilingual Knowledge Graph Construction (mKGC) refers to the task of automatically constructing or predicting missing entities and links for knowledge graphs in a multilingual setting.
In this work, we reformulate the mKGC task as a Question Answering (QA) task and introduce \texttt{mRAKL}: a Retrieval-Augmented Generation (RAG) based system to perform {\proj}. We achieve this by using the head entity and linking relation in a question, and having our model predict the tail entity as an answer.
Our experiments focus primarily on two low-resourced languages: Tigrinya and Amharic. We experiment with using higher-resourced languages Arabic and English for cross-lingual transfer. 
With a BM25 retriever, we find that the RAG-based approach improves performance over a no-context setting. Further, our ablation studies show that with an idealized retrieval system, \texttt{mRAKL} improves accuracy by 4.92 and 8.79 percentage points for Tigrinya and Amharic, respectively.
\end{abstract}

\section{Introduction}

Knowledge Graphs (KG) are structured multi-relational graphs that store factual knowledge. In a KG, nodes represent entities (e.g., Michelle Obama, Sasha Obama) and links represent relationships between the nodes (e.g., Michelle Obama - mother - Sasha Obama). Multilingual KGs are KGs in multiple languages. 

Despite their myriad downstream applications, including Question Answering~\cite{kgqa-2019-huang,UniKGQA-jiangunikgqa-2023}, Information Retrieval~\cite{kg-ir-2020-Reinanda}, and Language Model Augmentation~\cite{tian2024kg,wu2022efficient}, most KGs are incomplete~\cite{kgt5-saxena2022sequence,zhou-etal-2022-prix}. The quantity of missing information in KGs is even greater in low-resourced languages~\cite{zhou-etal-2022-prix}. Additionally, manual construction of KGs is expensive \cite{paulheim_2018}.  
 Recent work has investigated the use of pre-trained Language Models (LMs) for KG Construction~\cite[e.g.][]{kgt5-saxena2022sequence, yao2019kgbertbertknowledgegraph}. However, most of the work is focused on English, for which LMs have good performance~\cite{zhou-etal-2022-prix}. 

Multilingual Knowledge Graph Construction ({\proj}) research allows us to (1) extend the downstream benefits of KGs to multiple languages, and (2) capture culturally nuanced and relevant information across languages. However, the challenges of {\proj} are exacerbated for languages with limited data available. Prior work using LMs for {\proj} relies on pre-training LMs with large amounts of structured data (e.g., \citet{zhou-etal-2022-prix} train on a KG with 52M triples). However, languages on the long tail do not have such datasets available~\cite{joshi-etal-2020-state}.  Based on official statistics, only 0.2\% of the total entities in Wikidata~\cite{10.1145/2629489} have labels in the low-resourced language Amharic.\footnote{\url{https://www.wikidata.org/wiki/User:Pasleim/Language\_statistics\_for\_items}} Additionally, most pre-trained LMs do not have good performance for low-resourced languages~\cite{ojo2024goodlargelanguagemodels}. 

We propose \texttt{mRAKL}, a retrieval-augmented sequence-to-sequence generative method for {\proj}. \texttt{mRAKL} uses a \textit{retriever} which fetches relevant passages and passes them to a \textit{generator} model which predicts knowledge facts. 
Moreover, we allow LMs to learn better cross-lingual entity representation leveraging entity parallel textual information from Wikidata~\cite{10.1145/2629489}. These two approaches greatly alleviate the data scarcity problem. We focus on enriching KGs of two low-resourced languages: Amharic and Tigrinya, two Afro-Semitic languages.



\begin{figure*}[th]
    \centering
    \includegraphics[width=\linewidth]{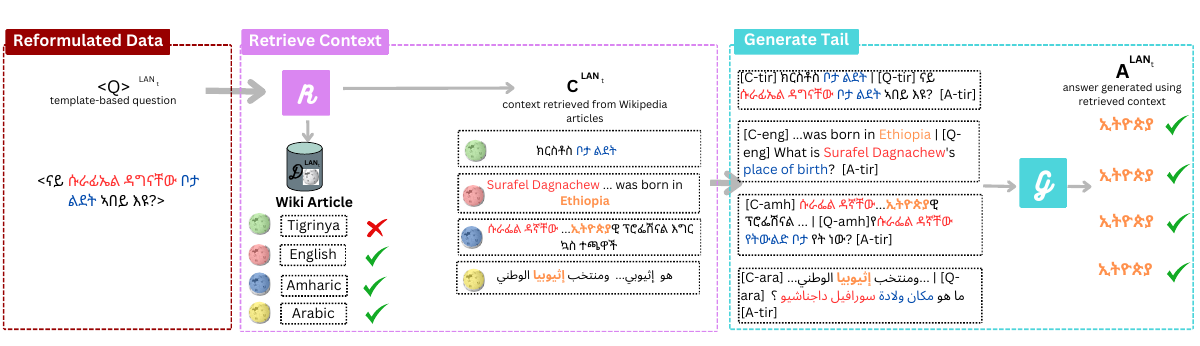}
    \caption{\textbf{Inference with \texttt{mRAKL} } In this example our triple is (\textcolor{red}{Surafel Dagnachew}, \textcolor{blue}{place of birth}, \textcolor{orange}{Ethiopia}). Taking the reformulated question ``What is \textcolor{red}{Surafel Dagnachew}'s \textcolor{blue}{place of birth}?'', the retriever encodes the query and fetches sentences with the highest similarity from the passages available from the Wikipedia articles in each of the languages. These sentences are then passed to the generator as context along with the question (see Appendix \ref{apn:data} for details on question generation in the four languages).
    }
    \label{fig:full_pipeline}
\end{figure*}

{\system} is a cross-lingual RAG-based QA system for {\proj} (see Figure \ref{fig:full_pipeline}). To use {\system}, we first reformulate {\proj} as a Question-Answering task. We create a QA dataset by transforming each KG triple into a question-answer pair: we construct a question text that uses the \textit{head} entity and \textit{relation} from the KG triple and an answer text that uses the \textit{tail} (see Figure \ref{fig:reformulation}). For our \textit{generator} model, we finetune a multilingual LM, AfriTeVa~\cite{jude-ogundepo-etal-2022-afriteva} with cross-lingual entity-centered alignment data. This finetuning allows the model to learn better representations for the entities in the low-resourced languages and unify complementary knowledge across languages.
Since data in these languages is small, we train a \textit{retriever} model to further enrich the construction by the \textit{generator}, utilizing monolingual datasets which are more easily available for low-resourced languages compared to structured, labeled datasets~\cite{joshi-etal-2020-state}

 We evaluate our approach and existing approaches on two tasks: probing parametric knowledge of pre-trained LMs for low-resourced languages (\textsection \ref{sec:parametric}) and monolingual and cross-lingual link prediction (\textsection \ref{sec:cross_link_prediction}).
 We also perform ablation studies showing the benefit of RAG for {\proj} where structured data is constrained (\textsection \ref{sec:additional}). Overall, we find that {\system} outperforms prior approaches on Amharic and Tigrinya by an 8 and 6 percentage point increase and that using cross-lingual context improves over no-context settings. Our contributions can be summarized as follows: 
 \begin{itemize}
     \item We contribute a 3.5k triple KG for Tigrinya and a 34k triple KG for Amharic. We also contribute our question templates and generative and retrieval models trained for {\proj} (\textsection \ref{sec:reformulation}).
     \footnote{We will release the link to our Github repo soon}
     \item We propose a novel RAG-based approach to {\proj} that retrieves relevant passages from unstructured monolingual data for KG completion (\textsection \ref{sec:training}). We show the benefits of using our method for low-resourced languages (\textsection \ref{sec:rag-link-prediction}).
     
     \item We propose a novel method for cross-lingual entity linking through cross-lingual link prediction, where given a \textit{head} entity and \textit{relation} in one language, the \textit{tail} is predicted in another language (\textsection \ref{sec:cross_link_prediction}). 
     
 \end{itemize}




\section{Related Work}


\paragraph{KG Construction}
Prior to LLMs, KGs were automatically constructed with multi-staged rule-based pipelines typically including information extraction, knowledge fusion, and knowledge graph completion~\cite[e.g.][]{lehmann2015dbpedia,hoffart2013yago2,dong2014knowledge}. Such systems are expensive to construct and maintain since they demand significant human efforts or only consume structured or semi-structured data which is easy for rule-based systems to deal with~\cite{carlson2010toward,vrandevcic2014wikidata,zhong2023comprehensive}. A recent trend explores ways to extract factual knowledge from LLMs with prompting and fine-tuning methods~\cite[e.g.][]{bosselut2019comet,jiang2020x,kgt5-saxena2022sequence,kassner2021multilingual,zhou-etal-2022-prix,song2023multilingual} both in monolingual and multilingual settings which addresses the drawback of traditional approaches. However, such approaches fall short in dealing with low-resourced languages when data is rarely seen in the training phase. {\system} employs RAG based approach to alleviate data scarcity. 


A number of multilingual KG embedding-based approaches have been proposed to tackle the cross-lingual knowledge alignment and KG completion problems~\cite[e.g][]{chen2021cross,chen2017multilingual,chakrabarti2022joint}. The key idea of these approaches is to align the knowledge across KGs in different languages into a unified embedding space so that the link prediction can leverage the complementary knowledge~\cite{chen-etal-2020-multilingual,sun2020knowledge}. Nevertheless, these approaches assume a closed-world framework which cannot take open-world information and natural language knowledge into consideration. Our system integrates the best of the worlds by combining knowledge from multilingual open-domain text through RAG and multilingual KGs through cross-entity alignment.

\paragraph{Retrieval Augmented Generation}
RAG augments LLMs with non-parametric memories from one or more external data sources. RAG has attracted significant attention recently because it addresses several critical limitations of LLMs~\cite[e.g.][]{guu2020realm,rag-llm-gao2023retrieval,rag-lewis2020retrieval}; for example, it is expensive to correct outdated, erroneous facts within LLMs through fine-tuning or re-training. RAG offers a way to circumvent this issue by providing up-to-date information to a pre-trained model during generation. 
Furthermore, RAG enables LLMs to focus on generalization and reasoning which reduces the size of LLMs needed to achieve similar performance~\cite{retrieval-lm-tutorial}.
Given that low-resource languages occupy the long tail of available knowledge during the training phase of LLMs, RAG offers a way to boost the performance for {\proj} by retrieving new, open-domain information instead of inferring knowledge only from multilingual KGs.
While several research works focus on augmenting LLMs with KGs~\cite{llm_kg,yang2024give}, to the best of our knowledge {\system} is the first work to explore how RAG with LLMs can help construct and complete KGs in a low-resourced setting.  




\section{Proposed Method} \label{sec:method}
Our proposed method is to reformulate {\proj} as a cross-lingual question-answering task and use an RAG-based QA system for completion. We first convert the \textit{(head, relation, tail)} triples into a \textit{<question, answer>} format where the question includes the \textit{head} and \textit{relation} and the answer is the \textit{tail} (\textsection \ref{sec:reformulation}). Using a \textit{retriever} model, we extract context from monolingual unlabeled datasets. We then use the \textit{generator} model to predict the tail entity, given the extracted context and the question (Figure \ref{fig:full_pipeline}). 
Below we first give background on our languages of focus, then detail our data preparation steps (\textsection \ref{sec:reformulation}) and then present {\system} (\textsection \ref{sec:training}).


\paragraph{Languages of Study}
Our main languages of study are Tigrinya and Amharic, Afro-Semitic languages that use the Ge'ez script. The languages are considered low-resourced in that there are limited tools and data available to build language technologies for them~\cite{yimam-etal-2020-exploring, gaim-etal-2023-question}. Tigrinya is one of the official languages of Ethiopia and Eritrea and is spoken by 9.7 million people\footnote{https://www.ethnologue.com/language/tir/} in total across the two countries and their diasporas. 
Amharic is one of the official languages of Ethiopia, spoken by over 33.7 million people as a first language and 25.1 million as a second language~\cite{10134103}. To enrich the data for these languages, we chose two transfer languages: Arabic and English. We selected Arabic as a transfer language because (1) it is in the same language family as Tigrinya and Amharic (motivated by \citet{ogunremi-etal-2023-mini})and (2) we hypothesize there are cultural and geographic ties that would make the information more culturally relevant (motivated by \citet{zhou-etal-2022-prix}). We select English as a second transfer language because of the abundance of resources in the language. 
\subsection{Data Preparation} \label{sec:reformulation}
In this section, we will describe our data collection process by detailing the steps we took for Tigrinya, one of our target languages. We will then give the statistics of the data for all of the languages in our experiments in Table \ref{tab:dataset}. The process when Amharic is the target language is exactly the same as detailed below for Tigrinya. When working on one of our target languages, we use the other languages as transfer languages; for instance, when Tigrinya is the target language, Amharic, English, and Arabic are transfer languages. 
\paragraph{Relations:} We first collected textual representations of the relations in Tigrinya from the Wikidata Property Explorer.\footnote{https://prop-explorer.toolforge.org/}
This resulted in 96 relations for Tigrinya. We then added 24 relations that have textual representations in Amharic but not in Tigrinya by manually translating the textual representations; this resulted in 120 relations in total.  
\paragraph{KG Extraction:} We extracted the Tigrinya KG from Wikidata using \textit{simple-wikidata-db.\footnote{https://github.com/neelguha/simple-wikidata-db/tree/main}} For each entity in Wikidata with a corresponding Wikipedia article title in the Tigrinya Wikipedia, we keep the triples from Wikidata that have the entity as a head or tail. We then filtered through the KG to keep the triples with the relations from our set of 120 relations described above. We extracted a KG with 3.5k edges and 272 unique entities. 
\paragraph{Template-Based Reformulation:} We then manually prepared question templates for each of the 120 relations\footnote{See Appendix \ref{apn:data} for translation and manual question preparation details.}.
Then, for each triple in the KG, we plug in the textual representation of the head entity into the template and use the textual representation of the tail entity as an answer, resulting in a 3.5k question-answer pair dataset. Figure \ref{fig:reformulation} shows an example of how we use our template-based reformulation approach. 
\paragraph{Extracting Context}
Given the \textit{head} and \textit{relation} in the question, the goal of the generator model is to predict the \textit{tail} as the answer. To enhance the ability of the \textit{generator} model in correctly predicting the tail, we use the \textit{retriever} model to extract sentences that will provide context. Since we do not have labeled data, we devised a heuristic to extract context for our question-answer pairs. We extracted the context for each question by searching for the tail entity in the first paragraph of the Tigrinya Wikipedia article associated with the head entity. We then kept a maximum of two sentences that had the tail entity as context. 
We use our heuristic context extraction method as an (im)perfect retriever setup: while it does not guarantee the context will always be retrieved (for example, when there is no mention of the tail entity in the head entity's Wikipedia article), when it does provide context, the retrieved context will certainly have the tail entity. However, this is not representative of the real-world task when we do not have access to the tail entity to search the Wikipedia article. 
Hence, we use the (im)perfect retriever setup to provide an upper bound of performance (\textsection \ref{sec:additional}).
\paragraph{Data in Transfer Languages:} We retrieved the textual representations for the 120 relations we collected as detailed above in Arabic and English from Wikidata using the relation ID. For Amharic, we manually translated the 68 relations that were unique to Tigrinya. Our final set of relations had 120 relations in the four languages of study. We then manually prepared template questions for each of the 120 relations in Amharic, English, and Arabic. Table \ref{tab:kg_size} gives details of the final dataset and Table \ref{tab:dataset} gives statistics on the coverage of each KG by the transfer languages.
We then got the labels for the head and tail entities in the Tigrinya KG from Wikidata in Amharic, English, and Arabic; extracting the 3.5k triples in our Tigrinya KG from the three transfer languages' KGs. Once we had the final dataset, we split it to train, evaluation and test sets with an 8:1:1 ratio. We use the evaluation set for hyperparameter tuning and report results on the test set which is unseen during training. We then used the same strategy as described for Tigrinya to perform the template-based reformulation and context extraction for Amharic (see Appendix \ref{apn:data}). 

  
\begin{table}[]
    \centering
    \begin{tabular}{cccc}
    \toprule
        KG  & Triples & Head & Tail \\
        \midrule
         Tigrinya & 3.5k & 244 & 170 \\
         Amharic & 34k & 8568 & 5058 \\
         \bottomrule
    \end{tabular}
    \caption{Details on size of KGs in the two target languages.}
    \label{tab:kg_size}
\end{table}
\begin{table}[]
    \centering
    \setlength\tabcolsep{2pt}
    \begin{tabular}{cp{1cm}cccc}
    \toprule
        & &   \multicolumn{2}{c|}{Tigrinya KG} & \multicolumn{2}{c}{Amharic KG} \\
        \cline{3-6}
        Language & Wiki& Head & Tail  & Head & Tail  \\
        \midrule
        Amharic & 14.04K& 79.50& 86.47  & 100 & 100 \\
        Arabic & 1.23M& 95.49 &99.41  & 79.56 & 94.36 \\
        English & 6.84M& 100&100  &90.40 & 98.39  \\
        Tigrinya & 506& 100& 100 & 3.60 & 4.03 \\
        \bottomrule
    \end{tabular}
    \caption{Percentage of the head and tail entities in each of the target language KGs with textual representations in each of the transfer languages.} 
    \label{tab:dataset}
\end{table}

\subsection{{\system}} \label{sec:training}
Our proposed setup involves a \textit{retriever} that extracts the necessary context and passes it to the \textit{generator} which, given context and a template question (i.e. a question with the relation and head entity), generates the answer (i.e. the tail). 
Figure \ref{fig:full_pipeline} shows our complete setup. 

\paragraph{Input Representation}  For what follows, we will use $t$ to refer to the target language. 
In the input sequence notation below, \texttt{LAN}\(_t\) represents the three-letter ISO language code for language \(t\). We use \texttt{Q} to represent the question text, \texttt{A} to represent the answer text, and \texttt{C} to represent the context text.  Our input sequences also use the special tokens \texttt[C-LAN], \texttt[Q-LAN], and \texttt[A-LAN] to indicate the start of a context, question, and answer respectively\footnote{This is inspired by prior work from \cite{zhou-etal-2022-prix}}. We use the `?' symbol to mark the end of a question\footnote{Note that we use \<؟> for Arabic questions.} and `|' to mark the end of a context. For the retriever, we pass the questions \texttt{Q} as queries and retrieve context \texttt{C} for each question. We concatenate the retrieved context \texttt{C} and the question \texttt{Q} as follows:

\begin{mdframed}
   \texttt{[C-\(LAN_{t}\)]C | [Q-\(LAN_{t}\)]Q? [A-\(LAN_{t}\)]} \\ $\forall$  \(t \epsilon \{Tigrinya, Amharic, English, Arabic\}\)
\end{mdframed}

The last element of the input sequence is the \texttt[A-LAN] token, which indicates in which language the model should generate answers. Hence, the retriever's task is, given \texttt{Q} with the \textit{head} and \textit{relation}, retrieve \texttt{C} that ideally includes the answer \texttt{A} which is the \textit{tail} entity. The generator's task is, given \texttt{C} which contains the \textit{tail} entity and \texttt{Q} with the \textit{head} and \textit{relation}, predict \texttt{A} which is the tail entity.

\paragraph{Training:} To train {\system} for link prediction, we prepare our template-based question-answering data as detailed in \textsection \ref{sec:reformulation}. For the triple (head, relation, tail), the question has the head and relation, and the answer is the tail (Figure \ref{fig:reformulation}). For the \textit{retriever} model, we use BM25 \cite{INR-019} and LaBSE\cite{feng-etal-2022-language}
For the \textit{generator} model, we finetune the AfriTeVa-base model with LoRA\cite{hu2022lora}.  During the \textit{generator} training, the model is trained with cross-entropy loss. Similar to \cite{saxena-etal-2022-sequence}, we do not use explicit negative sampling. We provide both \textit{generator} and \textit{retriever} model training details in Appendix \ref{apn:more_method_details}.

\paragraph{Inference:} Given a query (head, relation, ?), we first convert it to a question-answer format (\textsection \ref{sec:reformulation}). We then feed the question, which has the \textit{head} and \textit{relation}, to our \textit{retriever} to extract context from monolingual passages. We then feed the extracted context, which ideally would have the tail entity, to the \textit{generator}. The \textit{generator} takes the context and question as input and produces a probability distribution over all tokens. We use beam-search during decoding with a beam size of 10 and take the top n tokens (where n $\epsilon$ \{1,3,10\}) .

Given we are operating in a low-resourced context, we hypothesize that our RAG-based approach will improve performance over a generator-only approach by allowing the LM to learn the tail entity from an extracted context. Additionally, the multilingual setting allows for cross-lingual entity linking, i.e given a \textit{head} and \textit{relation} in one language, predicting the \textit{tail} in another language. This cross-lingual entity alignment enriches the dataset as well as the learned knowledge representation. Further, the modularity of the RAG pipeline allows us to improve the \textit{retriever} and \textit{generator} models separately; hence, we can utilize unlabeled, monolingual data which is easier to acquire than labeled and structured data \cite{joshi-etal-2020-state}, to perform {\proj}. 

\section{Experiments and Results}
\label{sec:exp}

In this section, we evaluate our proposed method with two tasks: (1) Parametric Knowledge Probing, which is a way to test knowledge representations in an LM's learned embeddings~(\textsection \ref{sec:parametric}) and (2) Link Prediction, which is a standard task for evaluating KG completion and construction. For Link Prediction, we evaluate by (1) comparing our method with that of prior work approaches (\textsection \ref{sec:rag-link-prediction}) and (2) by looking at cross-lingual link prediction, where we predict the tail entity in one language given a head and relation entity in another language  (\textsection \ref{sec:cross_link_prediction}). We also perform additional analysis with our (im)perfect retriever. 

Overall, we find that our approach outperforms prior methods by over a 4.9 percentage point increase and that providing context improves over a no-context setting by 6.7 and 12.22 percentage point increase. We also find that multilingual and cross-lingual approaches are more beneficial for Tigrinya, where the data is limited and the transfer languages cover the majority of the entities in the target language (see \textsection \ref{apn:data}). All results reported are based on a single inference run.

\subsection{Experimental Setup} \label{sec:experimental_setup}

Below, we describe the different experimental settings we tried for our retriever and generator.

\paragraph{Retriever:}  We use the Wikipedia articles from \cite{wikidump} and for all languages use the 2024-07-01 version \footnote{We access the data from https://huggingface.co/datasets/olm/wikipedia}. We experiment with the following retrieval setups:
\begin{itemize}
    \item \textbf{(Im)perfect Retriever:} As described in \textsection \ref{sec:reformulation}, we use our (im)perfect retriever to provide an approximate upper bound for how well our system will perform with a good retriever.
    \item \textbf{BM25}: We used the implementation by \citet{lù2024bm25sordersmagnitudefaster} for the BM25 indexes. We indexed the full Wikipedia articles of all head entities for retrieval. We built monolingual indexes for each of the four languages. (see Appendix \ref{apn:more_method_details} for details.)
    \item \textbf{LaBSE}: We use the LaBSE model~\cite{feng-etal-2022-language} as our retriever. LaBSE includes Amharic, Arabic, and English but not Tigrinya. We finetune the LaBSE model with contrastive loss by creating training dataset using our (im)perfect retriever (see Appendix \ref{apn:more_method_details} for details.)
    
\end{itemize}


\paragraph{Generator:} 
We experiment with four different setups for training the generator model. Since we are interested in performance both on Tigrinya and Amharic, we use each of the four setups for each of those two target languages.

\begin{itemize}
    \item No Context---question-answer pairs only without context. In this case, the input to our model is: \texttt{[Q-\(LAN_t\)]Q? [A-\(LAN_t\)]}.
    \item Monolingual Self-Context---where question-answer pairs have context in the target language only; the input to our model is: \texttt{[C-\(LAN_t\)]C |[Q-\(LAN_t\)]Q? [A-\(LAN_t\)]}.
    \item Multilingual Self-Context---where question-answer pairs along with the context are in the same language, for all four languages; the input to our model is: \texttt{[C-\(LAN_{t'}\)]C |[Q-\(LAN_{t'}\)]Q? [A-\(LAN_{t'}\)]} $\forall$ \(t' \epsilon \{Tigrinya, Amharic, English, Arabic\}\).
   \item Cross-Lingual Context---the context and the question are in the same language but the answer may be in any of the four languages; model input is: \texttt{[C-\(LAN_{t'}\)]C |[Q-\(LAN_{t'}\)]Q? [A-\(LAN_{t''}\)]} $\forall$  \(t', t'' \epsilon \{Tigrinya, Amharic, \\ English, Arabic\}\). 
    
\end{itemize} 

\begin{table}[]
    \centering
    \begin{tabular}{cccc} 
    \cline{3-4}
  \toprule
  Language $\rightarrow$ &   & Tigrinya & Amharic \\ 
    \midrule
    \multirow{3}{*}{Zero-Shot} &  mT5* & - & 0.49 \\
    & AfriTeVa $\dagger$ &0.22&0.61 \\
      & Aya* & 0.67 & 1.52\\
      & GPT-4& \textbf{2.23} & \textbf{5.83}\\
         \hline
    \multirow{2}{*}{Finetuned} & mT5 & 2.01 & 23.32 \\   
    & AfriTeva & \textbf{5.13} & \textbf{29.15} \\
       
         \bottomrule
    \end{tabular}
    \caption{\textbf{Zero-shot and finetuned model H@1 results for the no-context setup.} * indicates model does not include Tigrinya but includes Amharic. $\dagger$ indicates model includes both Tigrinya and Amharic. We omit the zero-shot performance of mT5 on Tigrinya as it was worse than AfriTeVa and the language is unseen for the model. 
    }
    \label{tab:parametric}
\end{table}

\subsection{Probing Parametric Knowledge} \label{sec:parametric}
\paragraph{Task Description} Probing parametric Knowledge of LMs involves using prompts to get predictions from a pre-trained LM\cite{petroni-etal-2019-language}. We perform this probing task in zero-shot on four pre-trained LMs. Testing the models in zero-shot gives us insight into the parametric knowledge \cite{yu-etal-2024-revealing} of the models for these languages. As our prompts, we use the template-based questions from our test set as described in \textsection \ref{sec:reformulation}.

\paragraph{Models in Comparison} We compare GPT-4o~\cite{openai2024gpt4technicalreport}, Aya\cite{ustun-etal-2024-aya}, mT5~\cite{xue-etal-2021-mt5}, and AfriTeVa~\cite{oladipo2023better}
. For mT5 and AfriTeva, we use the base models. AfriTeVa includes both target languages in its pre-training while Aya and mT5 include Amharic but not Tigrinya. Arabic and English are included in all models. 

\paragraph{Metric:} We use H@1 as our metric. We count it as a hit if the prediction contains the tail entity; for instance, if the target is ``Addis Ababa'' and the model prediction is ``It is Addis Ababa,'' we count it as a hit\footnote{For details on how we attempted to constrain model outputs to predict the tail entity only, refer to Appendix \ref{apn:data}.}.

\paragraph{Results and Discussion} All models perform poorly in both languages in a zero-shot setting, never surpassing 6\% (Table \ref{tab:parametric}). We find that the large, generative models, GPT-4o and Aya, outperform the Seq2Seq models with GPT-4o having the highest performance for zero-shot.
For Tigrinya, the mT5 model did not produce meaningful predictions in the target language; we hypothesize this is due to the language being unseen for the model and omit the results from the table.   
We find that AfriTeVa outperforms mT5 on our dataset both before and after fine-tuning. Based on these results, we use AfriTeVa as our base model.

\subsection{Closed vs Open Domain Link Prediction} \label{sec:rag-link-prediction}
\paragraph{Task Description} Link prediction in KG Construction literature \cite{zhou-etal-2022-prix} refers to the task of predicting a tail entity given a head entity and relation. 
In this section, we study the impact of RAG for {\system}. For a baseline comparison, we adopt the setting of KGT5~\cite{kgt5-saxena2022sequence} and KGT5-context~\cite{kochsiek-etal-2023-friendly}. KGT5 is a monolingual model trained on the Wikidata5M~\cite{wang-etal-2021-kepler} dataset. We use their verbalization scheme for link prediction: given a triple (\textit{head, relation, tail}), the input to the model is \texttt{``predict tail: \textit{head} | \textit{relation}''} and the expected output is the tail entity. We train our base model, AfriTeVa-base, with this scheme on our dataset and compare it to our No-Context setup. For comparison with our RAG-based system, we adopt the setup from \citet{kochsiek-etal-2023-friendly} where the model is given additional context during training by (1) appending the description of each entity from Wikipedia (KGT5-Description) and (2) appending the entities directly connected to the head entity (KGT5-One-Hop). 

\begin{table}[] 
    \centering
    
    \scalebox{1}{
    \begin{tabular}{cp{0.6cm}p{0.6cm}p{0.6cm}p{0.6cm}} 
    \toprule
      & \multicolumn{2}{c}{Tigrinya KG} & \multicolumn{2}{c}{Amharic KG} \\ 
    \cline{2-5}
     & H@1 & H@10  & H@1 & H@10\\ 
      \midrule
   
 KGT5-No-Context & 6.91  & 28.57  & 32.58  & 52.57\\
    KGT5-Description & 5.8 & 23.44 & 32.91 & 43.32 \\
    KGT5-One-Hop & 4.46 & 24.33 & 28.83& 48.17\\
    \midrule
    (ours) No-Context & 5.13 & 26.11 & 29.15& 54.81\\
      (ours) Self-Context & \textbf{11.83} & \textbf{34.59} & \textbf{41.37}& \textbf{61.87}\\
    \bottomrule
    \end{tabular}
    }\caption{Comparison of our proposed method with that of prior work for low-resourced languages where there is limited structured data in a Monolingual setting.
    }
    \label{tab:context_is_better}
\end{table}
\paragraph{Metric:} We use H@1, H@3, and H@10 to denote Hit at the models' top 1, top 3, and top 10 predictions. Hit is counted only if the prediction is an Exact Match (EM) of the tail entity.

\paragraph{Results and Discussion} As Table \ref{tab:context_is_better} shows, we find that our proposed method with context outperforms the adopted approach from prior work by a 4.92 and 8.79 increase in percentage points for Tigrinya and Amharic respectively for H@1\footnote{Performance gain calculated by taking the difference between best-performing methods for each work.}. While in the No-context case, the KGT5 approach outperforms our no-context setting by 1.78 and 3.43 percentage points, we see that adding the structured context (i.e descriptions of entities from Wikidata and one-hop connections) degrades the performance for our target languages. Table \ref{tab:kgt5_reason} shows that structured context required by prior work is not readily avaiable for these languages and when it is avaiable, it rarely contains the tail entity. We hypothesize this limited availability of descriptions and one-hop connections in the low-resourced languages leads to closed-domain link prediction methods being limited for low-resourced languages. 
Hence, we find that our approach of using unstructured data for retrieving context is a better approach for low-resourced languages.

\begin{table*}[]
    \centering
    \setlength\tabcolsep{3pt}
    \begin{tabular}{cccccccccccc}
    \toprule
 Target lang. $\rightarrow$ & &  \multicolumn{5}{c}{Tigrinya} & \multicolumn{5}{c}{Amharic}\\
    \hline
   Context lang. $\rightarrow$ &   & Amh& Ara& Eng& Tir& Avg. $\mid$   & Amh& Ara& Eng& Tir& Avg. \\
    \midrule
   \multirow{3}{*}{H@1} 
   & No-Context  & 11.64 & 12.08 & 14.06 & 14.06 & 12.97  $\mid$ & 35.89  & 31.51 &36.63 & 8.29& 33.12\\ 
 
   & LaBSE  & 12.10 & 10.29 &13.17 &13.62 &12.30  $\mid$ &34.27 & 30.29 &36.07 &10.49 &32.19 \\
    &  BM25 & \textbf{15.75} & \textbf{12.30} & \textbf{14.73} & \textbf{13.84} & \textbf{14.15} $\mid$ &  \textbf{38.52} &\textbf{33.58} &\textbf{38.22 }& \textbf{11.17}&\textbf{35.27}  \\ 
    \midrule
  \multirow{3}{*}{H@3} 
  &  No-Context  & \textbf{22.60} & \textbf{21.48} & 22.77 & 22.32 & 22.29 $\mid$  & 44.70& 40.41& 45.69& \textbf{17.60}& 42.06\\ 
  
  & LaBSE  & 21.19&18.12 &20.76  &20.38 &22.53 $\mid$    &43.80&39.16 &44.76 &17.26 &41.08 \\
   & BM25  & 21.92& 21.70 &\textbf{23.21} &\textbf{23.44} &\textbf{22.57}  $\mid$  & \textbf{46.32}& \textbf{41.34}& \textbf{46.62}&16.75 &\textbf{43.11} \\ 
  
    \midrule
      \multirow{3}{*}{H@10} 
  &  No-Context  & \textbf{39.72}& 36.91&38.83 &38.16 &38.40  $\mid$ & 54.65 & 49.86&\textbf{56.04} &29.95 &52.14 \\ 
   & LaBSE &39.50 &36.02 &36.38 &37.95 &37.45  $\mid$  & 52.48  & 48.28& 53.52& \textbf{30.80} &50.22 \\
    & BM25  & 37.67& \textbf{35.35} &\textbf{38.17} &\textbf{38.62} &\textbf{37.45}  $\mid$ &\textbf{54.97} &\textbf{50.25} &55.44 &30.12 &\textbf{52.18} \\

    \bottomrule
    \end{tabular}
    \caption{Cross-Lingual Link Prediction results broken down by the context language. }
    \label{tab:bm25_performance}
\end{table*}

\subsection{Cross-Lingual Link Prediction} \label{sec:cross_link_prediction}
\paragraph{Task Description}  In addition to the monolingual link prediction task in Section~\ref{sec:rag-link-prediction}, we propose and perform a cross-lingual link prediction task where the head and relation are in one language and the tail is in another language (see \textsection \ref{sec:experimental_setup}). Here, we do not compare to KGT5 or KGT5-Context as the approach is for monolingual settings. Instead, we compare two retrievers and a no-context setting.



    
  

\paragraph{Results and Discussion}
Table \ref{tab:bm25_performance} shows performance for the cross-lingual link prediction task. We find that using the LaBSE retriever does not improve performance over the BM25 retriever. However, both retriever options show a gain in performance as compared to the no-context setup.  We also observe that the Hit@10 with Amharic as a context is highest compared with other languages as context depending on the available context and the overlap with the target language. We provide detailed analysis in Section~\ref{sec:additional}.

\subsection{Additional Analysis} \label{sec:additional}
\paragraph{Effects of Multilingual Context} As Table \ref{tab:multi_vs_mono} shows, with the (im)perfect retriever, Multilingual Self-Context---using multiple languages but matching the language of any given prompt's context, question, and answer---improves performance over training only with the target language data. The performance boost is especially pronounced for Tigrinya, where adding context results in a 4.69 percentage point increase compared to the Monolingual Self-Context setting. As Figure \ref{fig:tigrinya_breakdown} shows, for the Tigrinya KG, we observe that Arabic and English each provide context for ~25\% of the dataset while the Amharic and Tigrinya each provide context for less than 10\% of the test data. However, we see that H@1 is the highest when Amharic provides context. As demonstrated in Table \ref{tab:dataset}, 86.47\% of the tail entities in Tigrinya have corresponding labels in Amharic. We further investigated the overlap of tail entities between the two languages that are spelled the same--i.e would have the same learned representation in the model--and found that 35.88\% of the tail entities in Amharic and Tigrinya share the same spelling (Figure \ref{fig:shared_entities}). This partially explains the boost we observe for using Amharic context.
Similarly, for Amharic KG, Figure \ref{fig:amharic_breakdown} shows that Arabic provides context for ~25\% of the dataset while English provides context for ~30\%. Tigrinya provides context for less than 1\% of the data. Amharic, the target language, provides context for ~12\% of the test set. We see that performance is highest when Amharic provides context. In both Amharic and Tigrinya KGs, while questions without context dominate the test set (43\% and 34\% respectively), performance on those questions is the lowest, showing the advantage of cross-lingual context.

\begin{table}[] 
    \centering
    \setlength\tabcolsep{3pt}
    \scalebox{1}{
    \begin{tabular}{p{1cm}ccc|ccc} 
    \toprule
     & \multicolumn{3}{c}{Tigrinya} & \multicolumn{3}{c}{Amharic} \\ 
    \cline{2-7}
    
     & None & Target  & Avg.  & None & Target & Avg.\\ 
      \midrule
      Mono & 7.34 & 54.76 & 11.83 & \textbf{30.70} & 79.47 & 42.92 \\
      \hline
      Multi &  \textbf{9.60} & \textbf{69.05}  &15.18  & 30.53 & \textbf{80.85} & \textbf{43.21} \\
      
      \bottomrule
    \end{tabular}
    }\caption{Breakdown of H@1 results by the availability of context in the target language with the (im)perfect retriever. Results for Monolingual-Self Context (Mono) and Multilingual-Self Context (Multi) settings. 
    }
    \label{tab:multi_vs_mono}
\end{table}


\paragraph{Qualitative Analysis} We qualitatively looked at model predictions to get insights into what the models learned. In Figure \ref{fig:multilingual}, we show an example from the BM25 retriever where the tail entity is in the retrieved context for Arabic but not in the context for Amharic (our target language) or English. The Amharic context is unrelated to the query, ``What is Blue an instance of?''; the English context is related in that it talks about a blue dye, even though it does not use the term `color'; the Arabic context talks about the blue, red, and green colors of the Azerbaijan flag, providing context that our head entity ``blue'' is a type of color. In this example, we see that our system benefits from the cross-lingual transfer, where the entity names in the three languages are aligned and the model can correctly predict the head entity in all three languages. This is an instance of how our cross-lingual entity alignment (Sec. \ref{sec:cross_link_prediction}) works. While prior work \cite[e.g.][]{zhou-etal-2022-prix} relies on explicitly aligning entities in different languages, in our approach, the alignment is done through the cross-lingual context provided to the generative model. Refer to Appendix \ref{apn:more_results} for additional qualitative analysis. 

Further, we looked at cases where the transfer language is English as compared to Arabic. Supporting our hypothesis, we find that Arabic helps with some  queries that are regionally specific to the target languages: for instance, queries like ``What is Afar’s writing script?'' where the English provided context outputs an incorrect prediction. Further, we observe that the Arabic context helps for queries related to Middle Eastern and Asian contexts such as ``What is Arwad’s country?'' and ``What is Gwadar Port's country?'' On the other hand, queries like ``What country is Madrid the capital of?'' were correctly predicted as ``Spain'' when English context was provided, but ``Afghanistan'' when Arabic context was provided. This suggests that the English context provides support for more Western topics while the Arabic context provides support for more culturally/regionally specific and Middle East/Asian queries. Hence, there is potential for future work to explore how to best select transfer languages that support diverse cultures.


\paragraph{Bias in Data} Prior work has demonstrated that there are societal biases in Wikipedia across languages~\cite{samir-etal-2024-locating}. Specifically in relation to our languages of study, prior work has shown that Wikipedia articles might contain ``harmful and abusive content.''\cite{10.1145/3613904.3642605}. In light of these prior works, it is crucial to interrogate our findings and data. In this work, we specifically looked at gaps between the KGs of each of the target languages. We find that entities that exist in Tigrinya KG but not in Amharic KG are mostly regions in Eritrea like the Northern Red Sea Region and Gash-Barka Region. Hence, using Amharic as a transfer language provides data for entities shared in common by Ethiopia and Eritrea but lacks representation for entities that are exclusively related to Eritrea. Similarly, we find that what is not covered by the transfer languages for the Amharic KG is mostly related to famous Ethiopians. Therefore, while transfer languages can help provide context for entities that are shared across the languages, there are gaps in transfer language KGs for entities that are specific to the target languages. We give more details on this in Appendix \ref{apn:data}.

\begin{figure}
    \centering
    \includegraphics[width=\linewidth]{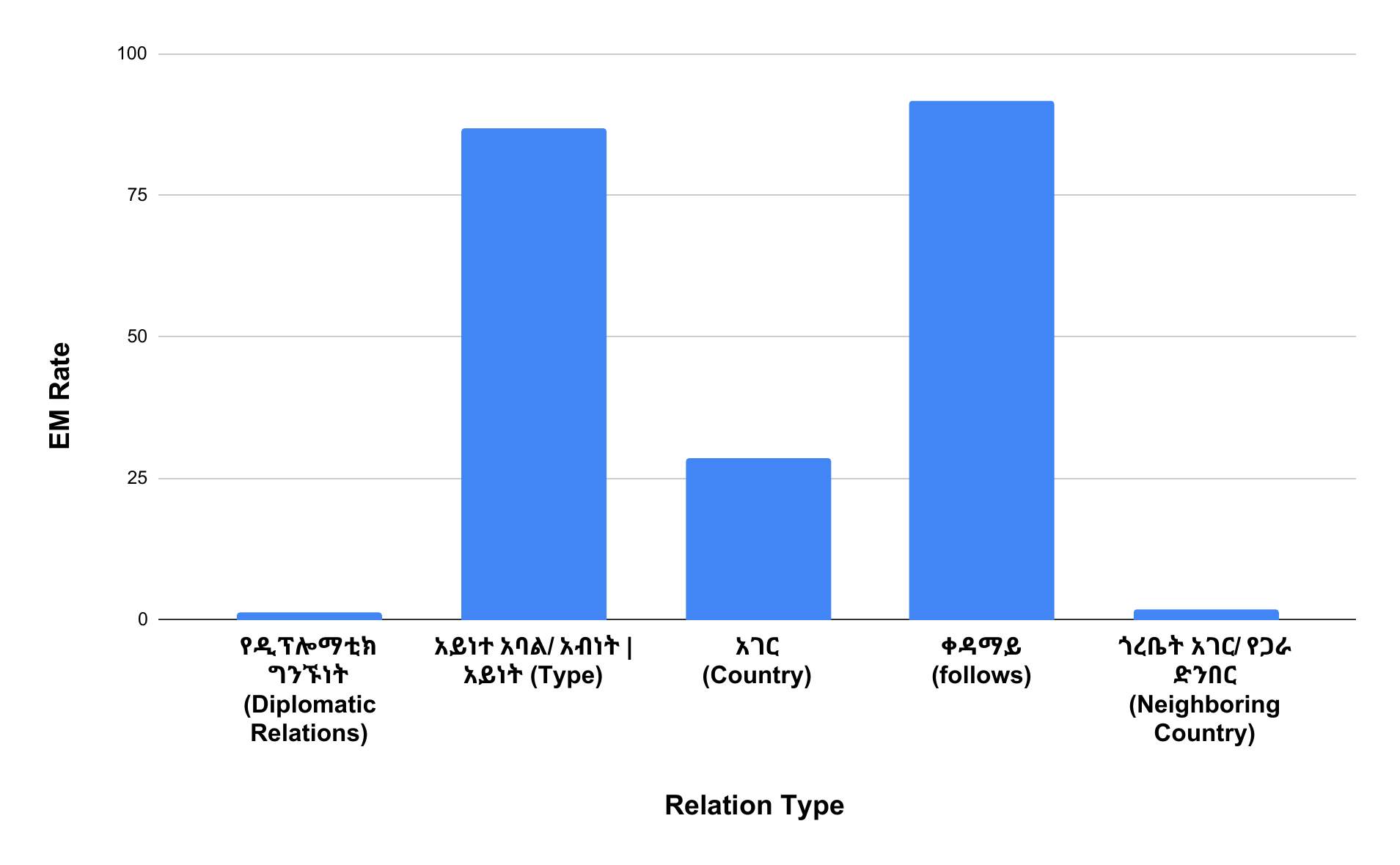}
    \caption{Percentage of triples with the given relation that had correct tail predictions for top 5 most frequent relations in Amharic test set.}
    \label{fig:relations_amh}
\end{figure}
\begin{figure}
    \centering
  \includegraphics[width=\linewidth]{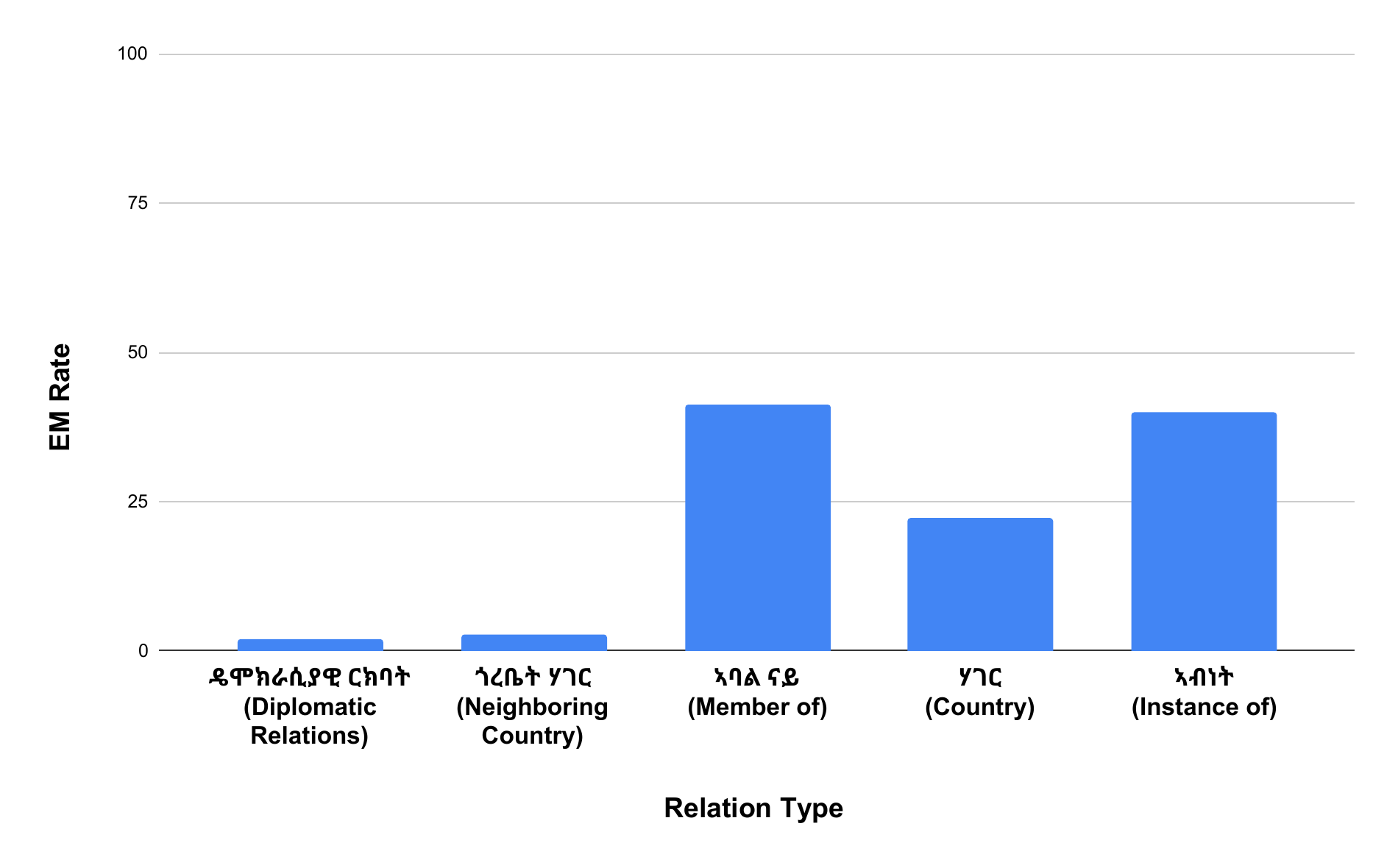}
    \caption{Percentage of triples with the given relation that had correct tail predictions for top 5 most frequent relations in Tigrinya test set.}
    \label{fig:relations_tir}
\end{figure}
\paragraph{Analysis by Relation Type} To understand which relations were being correctly linked, we looked at the triples for which our system correctly predicted the tail. Specifically, we looked at the distribution of the relations for which the triples were correctly completed. We used the BM25 retriever model setup for this analysis. For both target languages, ``diplomatic relations'' is the most frequent relation type; i.e majority of the triples in the test set have the relation ``diplomatic relations.''. In Figures \ref{fig:relations_amh} and \ref{fig:relations_tir}, we show the distribution of the top 5 most frequent relations in the Amharic and Tigrinya test set respectively, along with the percentage of triples with each relation that were correctly predicted by the mRAKL system. In both target language cases, triples with ``diplomatic relations'' and ``neighboring country'' have the least percentage of correctly predicted tails. This could be because they have a many-to-many relationship. On the other hand, we observe that Amharic has higher percentage of triples with top 5 most frequent relations correctly predicted, indicating the benefits of training from more data as the model will have access to additional context.

\section{Conclusion}
We propose~{\system}, a RAG-based approach for {\proj} in low-resourced language settings. We have shown that a RAG-based finetuned LM can retrieve facts that help with {\proj}. In addition, our cross-lingual entity alignment technique combines complementary knowledge across languages and increases the available corpus for RAG.
Our experimental results demonstrate that {\system} increases accuracy by up to $4.92$ and $8.79$ percentage points for Tigrinya and Amharic, respectively, compared with baselines. Our approach represents a step towards alleviating the cultural knowledge scarcity that LLMs typically display during pre-training.  

\section*{Limitations}
Our work has several limitations: First, when using the (im)perfect retriever, the extracted context may not be the exact context needed (see \textsection \ref{sec:method} and Appendix \ref{apn:more_method_details}). However, it does correctly link the head and relation to the tail entity. With the limited labeled data available for our languages of focus, the heuristic method was the best option we could come up with to provide an upper bound on what {\system} can achieve. Additionally, we focused our efforts on two low-resourced languages. We do not make claims about the efficacy of our method for other low-resourced languages; we only had resources (human and computational) sufficient to work on the two languages. However, low-resourced languages differ in the available resources; some languages may have even more limited monolingual data. To account for this, we provide results with and without context. Additionally, the two languages are related and use the same script. Future work can explore to what extent our approach can be extended to other low-resourced languages. We also relied on manual effort to construct the templates for each of the languages. While this requires human labor, we were interested in providing high-quality data for these languages. Future work could explore using automated methods (e.g using machine translation). Additionally, our work does not look into co-reference resolution, i.e, resolving multiple names of an entity. Currently, our work relies on the generative model to implicitly resolve multiple entity names. We will explore more explicit co-reference resolution techniques in future work.

\section*{Acknowledgments}
We would like to thank members and friends of PLAIT, EPIC and Canny Lab for their feedback on this work. We also want to thank Nuredin Ali and Negasi Haile for their help in reviewing the template questions for Tigrinya and Mustafa Abuzahriyeh, Habiba Geweifal, and Anisa Mohammed for their help with the Arabic template questions. We thank Hailay Teklehaymanot for early conversations about this work. We thank the reviewers for their valuable feedback on our paper.

\bibliography{anthology,custom}

\begin{thebibliography}{55}
\expandafter\ifx\csname natexlab\endcsname\relax\def\natexlab#1{#1}\fi

\bibitem[{Asai et~al.(2023)Asai, Min, Zhong, and Chen}]{retrieval-lm-tutorial}
Akari Asai, Sewon Min, Zexuan Zhong, and Danqi Chen. 2023.
\newblock Acl 2023 tutorial: Retrieval-based language models and applications.
\newblock \emph{ACL 2023}.

\bibitem[{Basha et~al.(2023)Basha, Veeraiah, Charan, Joyce~Yeddu, and Babu}]{10134103}
Shaik~Johny Basha, Duggineni Veeraiah, Boddu~Venkat Charan, Wiltrud~Sahithi Joyce~Yeddu, and Devalla~Ganesh Babu. 2023.
\newblock \href {https://doi.org/10.1109/ICICT57646.2023.10134103} {Detection and comparative analysis of handwritten words of amharic language to english using cnn-based frameworks}.
\newblock In \emph{2023 International Conference on Inventive Computation Technologies (ICICT)}, pages 422--427.

\bibitem[{Bosselut et~al.(2019)Bosselut, Rashkin, Sap, Malaviya, Celikyilmaz, and Choi}]{bosselut2019comet}
Antoine Bosselut, Hannah Rashkin, Maarten Sap, Chaitanya Malaviya, Asli Celikyilmaz, and Yejin Choi. 2019.
\newblock Comet: Commonsense transformers for automatic knowledge graph construction.
\newblock In \emph{Proceedings of the 57th Annual Meeting of the Association for Computational Linguistics}, pages 4762--4779.

\bibitem[{Carlson et~al.(2010)Carlson, Betteridge, Kisiel, Settles, Hruschka, and Mitchell}]{carlson2010toward}
Andrew Carlson, Justin Betteridge, Bryan Kisiel, Burr Settles, Estevam Hruschka, and Tom Mitchell. 2010.
\newblock Toward an architecture for never-ending language learning.
\newblock In \emph{Proceedings of the AAAI conference on artificial intelligence}, volume~24, pages 1306--1313.

\bibitem[{Chakrabarti et~al.(2022)Chakrabarti, Singh, Lohiya, Jain et~al.}]{chakrabarti2022joint}
Soumen Chakrabarti, Harkanwar Singh, Shubham Lohiya, Prachi Jain, et~al. 2022.
\newblock Joint completion and alignment of multilingual knowledge graphs.
\newblock In \emph{Proceedings of the 2022 Conference on Empirical Methods in Natural Language Processing}, page 11922—11938.

\bibitem[{Chen et~al.(2021)Chen, Shi, Zhou, and Roth}]{chen2021cross}
Muhao Chen, Weijia Shi, Ben Zhou, and Dan Roth. 2021.
\newblock Cross-lingual entity alignment with incidental supervision.
\newblock In \emph{Proceedings of the 16th Conference of the European Chapter of the Association for Computational Linguistics: Main Volume}, pages 645--658.

\bibitem[{Chen et~al.(2017)Chen, Tian, Yang, and Zaniolo}]{chen2017multilingual}
Muhao Chen, Yingtao Tian, Mohan Yang, and Carlo Zaniolo. 2017.
\newblock Multilingual knowledge graph embeddings for cross-lingual knowledge alignment.
\newblock In \emph{Proceedings of the Twenty-Sixth International Joint Conference on Artificial Intelligence}. International Joint Conferences on Artificial Intelligence Organization.

\bibitem[{Chen et~al.(2020)Chen, Chen, Fan, Uppunda, Sun, and Zaniolo}]{chen-etal-2020-multilingual}
Xuelu Chen, Muhao Chen, Changjun Fan, Ankith Uppunda, Yizhou Sun, and Carlo Zaniolo. 2020.
\newblock \href {https://doi.org/10.18653/v1/2020.findings-emnlp.290} {Multilingual knowledge graph completion via ensemble knowledge transfer}.
\newblock In \emph{Findings of the Association for Computational Linguistics: EMNLP 2020}, pages 3227--3238, Online. Association for Computational Linguistics.

\bibitem[{Dong et~al.(2014)Dong, Gabrilovich, Heitz, Horn, Lao, Murphy, Strohmann, Sun, and Zhang}]{dong2014knowledge}
Xin Dong, Evgeniy Gabrilovich, Geremy Heitz, Wilko Horn, Ni~Lao, Kevin Murphy, Thomas Strohmann, Shaohua Sun, and Wei Zhang. 2014.
\newblock Knowledge vault: A web-scale approach to probabilistic knowledge fusion.
\newblock In \emph{Proceedings of the 20th ACM SIGKDD international conference on Knowledge discovery and data mining}, pages 601--610.

\bibitem[{Feng et~al.(2022)Feng, Yang, Cer, Arivazhagan, and Wang}]{feng-etal-2022-language}
Fangxiaoyu Feng, Yinfei Yang, Daniel Cer, Naveen Arivazhagan, and Wei Wang. 2022.
\newblock \href {https://doi.org/10.18653/v1/2022.acl-long.62} {Language-agnostic {BERT} sentence embedding}.
\newblock In \emph{Proceedings of the 60th Annual Meeting of the Association for Computational Linguistics (Volume 1: Long Papers)}, pages 878--891, Dublin, Ireland. Association for Computational Linguistics.

\bibitem[{Foundation()}]{wikidump}
Wikimedia Foundation.
\newblock \href {https://dumps.wikimedia.org} {Wikimedia downloads}.

\bibitem[{Gaim et~al.(2023)Gaim, Yang, Park, and Park}]{gaim-etal-2023-question}
Fitsum Gaim, Wonsuk Yang, Hancheol Park, and Jong Park. 2023.
\newblock \href {https://doi.org/10.18653/v1/2023.acl-long.661} {Question-answering in a low-resourced language: Benchmark dataset and models for {T}igrinya}.
\newblock In \emph{Proceedings of the 61st Annual Meeting of the Association for Computational Linguistics (Volume 1: Long Papers)}, pages 11857--11870, Toronto, Canada. Association for Computational Linguistics.

\bibitem[{Gao et~al.(2023)Gao, Xiong, Gao, Jia, Pan, Bi, Dai, Sun, and Wang}]{rag-llm-gao2023retrieval}
Yunfan Gao, Yun Xiong, Xinyu Gao, Kangxiang Jia, Jinliu Pan, Yuxi Bi, Yi~Dai, Jiawei Sun, and Haofen Wang. 2023.
\newblock Retrieval-augmented generation for large language models: A survey.
\newblock \emph{arXiv preprint arXiv:2312.10997}.

\bibitem[{Guu et~al.(2020)Guu, Lee, Tung, Pasupat, and Chang}]{guu2020realm}
Kelvin Guu, Kenton Lee, Zora Tung, Panupong Pasupat, and Ming-Wei Chang. 2020.
\newblock Realm: retrieval-augmented language model pre-training.
\newblock In \emph{Proceedings of the 37th International Conference on Machine Learning}, pages 3929--3938.

\bibitem[{Hoffart et~al.(2013)Hoffart, Suchanek, Berberich, and Weikum}]{hoffart2013yago2}
Johannes Hoffart, Fabian~M Suchanek, Klaus Berberich, and Gerhard Weikum. 2013.
\newblock Yago2: A spatially and temporally enhanced knowledge base from wikipedia.
\newblock \emph{Artificial intelligence}, 194:28--61.

\bibitem[{Hu et~al.(2022)Hu, yelong shen, Wallis, Allen-Zhu, Li, Wang, Wang, and Chen}]{hu2022lora}
Edward~J Hu, yelong shen, Phillip Wallis, Zeyuan Allen-Zhu, Yuanzhi Li, Shean Wang, Lu~Wang, and Weizhu Chen. 2022.
\newblock \href {https://openreview.net/forum?id=nZeVKeeFYf9} {Lo{RA}: Low-rank adaptation of large language models}.
\newblock In \emph{International Conference on Learning Representations}.

\bibitem[{Huang et~al.(2019)Huang, Zhang, Li, and Li}]{kgqa-2019-huang}
Xiao Huang, Jingyuan Zhang, Dingcheng Li, and Ping Li. 2019.
\newblock \href {https://doi.org/10.1145/3289600.3290956} {Knowledge graph embedding based question answering}.
\newblock In \emph{Proceedings of the Twelfth ACM International Conference on Web Search and Data Mining}, WSDM '19, page 105–113, New York, NY, USA. Association for Computing Machinery.

\bibitem[{Jiang et~al.(2023)Jiang, Zhou, Zhao, and Wen}]{UniKGQA-jiangunikgqa-2023}
Jinhao Jiang, Kun Zhou, Xin Zhao, and Ji-Rong Wen. 2023.
\newblock Unikgqa: Unified retrieval and reasoning for solving multi-hop question answering over knowledge graph.
\newblock In \emph{The Eleventh International Conference on Learning Representations}.

\bibitem[{Jiang et~al.(2020)Jiang, Anastasopoulos, Araki, Ding, and Neubig}]{jiang2020x}
Zhengbao Jiang, Antonios Anastasopoulos, Jun Araki, Haibo Ding, and Graham Neubig. 2020.
\newblock X-factr: Multilingual factual knowledge retrieval from pretrained language models.
\newblock In \emph{Proceedings of the 2020 Conference on Empirical Methods in Natural Language Processing (EMNLP)}, pages 5943--5959.

\bibitem[{Joshi et~al.(2020)Joshi, Santy, Budhiraja, Bali, and Choudhury}]{joshi-etal-2020-state}
Pratik Joshi, Sebastin Santy, Amar Budhiraja, Kalika Bali, and Monojit Choudhury. 2020.
\newblock \href {https://doi.org/10.18653/v1/2020.acl-main.560} {The state and fate of linguistic diversity and inclusion in the {NLP} world}.
\newblock In \emph{Proceedings of the 58th Annual Meeting of the Association for Computational Linguistics}, pages 6282--6293, Online. Association for Computational Linguistics.

\bibitem[{Jude~Ogundepo et~al.(2022)Jude~Ogundepo, Oladipo, Adeyemi, Ogueji, and Lin}]{jude-ogundepo-etal-2022-afriteva}
Odunayo Jude~Ogundepo, Akintunde Oladipo, Mofetoluwa Adeyemi, Kelechi Ogueji, and Jimmy Lin. 2022.
\newblock \href {https://doi.org/10.18653/v1/2022.deeplo-1.14} {{A}fri{T}e{VA}: Extending ?small data? pretraining approaches to sequence-to-sequence models}.
\newblock In \emph{Proceedings of the Third Workshop on Deep Learning for Low-Resource Natural Language Processing}, pages 126--135, Hybrid. Association for Computational Linguistics.

\bibitem[{Kassner et~al.(2021)Kassner, Dufter, and Sch{\"u}tze}]{kassner2021multilingual}
Nora Kassner, Philipp Dufter, and Hinrich Sch{\"u}tze. 2021.
\newblock Multilingual lama: Investigating knowledge in multilingual pretrained language models.
\newblock In \emph{Proceedings of the 16th Conference of the European Chapter of the Association for Computational Linguistics: Main Volume}, pages 3250--3258.

\bibitem[{Kochsiek et~al.(2023)Kochsiek, Saxena, Nair, and Gemulla}]{kochsiek-etal-2023-friendly}
Adrian Kochsiek, Apoorv Saxena, Inderjeet Nair, and Rainer Gemulla. 2023.
\newblock \href {https://doi.org/10.18653/v1/2023.repl4nlp-1.11} {Friendly neighbors: Contextualized sequence-to-sequence link prediction}.
\newblock In \emph{Proceedings of the 8th Workshop on Representation Learning for NLP (RepL4NLP 2023)}, pages 131--138, Toronto, Canada. Association for Computational Linguistics.

\bibitem[{Lehmann et~al.(2015)Lehmann, Isele, Jakob, Jentzsch, Kontokostas, Mendes, Hellmann, Morsey, Van~Kleef, Auer et~al.}]{lehmann2015dbpedia}
Jens Lehmann, Robert Isele, Max Jakob, Anja Jentzsch, Dimitris Kontokostas, Pablo~N Mendes, Sebastian Hellmann, Mohamed Morsey, Patrick Van~Kleef, S{\"o}ren Auer, et~al. 2015.
\newblock Dbpedia--a large-scale, multilingual knowledge base extracted from wikipedia.
\newblock \emph{Semantic web}, 6(2):167--195.

\bibitem[{Lewis et~al.(2020)Lewis, Perez, Piktus, Petroni, Karpukhin, Goyal, K{\"u}ttler, Lewis, Yih, Rockt{\"a}schel et~al.}]{rag-lewis2020retrieval}
Patrick Lewis, Ethan Perez, Aleksandra Piktus, Fabio Petroni, Vladimir Karpukhin, Naman Goyal, Heinrich K{\"u}ttler, Mike Lewis, Wen-tau Yih, Tim Rockt{\"a}schel, et~al. 2020.
\newblock Retrieval-augmented generation for knowledge-intensive nlp tasks.
\newblock \emph{Advances in Neural Information Processing Systems}, 33:9459--9474.

\bibitem[{Lù(2024)}]{lù2024bm25sordersmagnitudefaster}
Xing~Han Lù. 2024.
\newblock \href {http://arxiv.org/abs/2407.03618} {Bm25s: Orders of magnitude faster lexical search via eager sparse scoring}.

\bibitem[{Nigatu et~al.(2024)Nigatu, Canny, and Chasins}]{10.1145/3613904.3642605}
Hellina~Hailu Nigatu, John Canny, and Sarah~E. Chasins. 2024.
\newblock \href {https://doi.org/10.1145/3613904.3642605} {Low-resourced languages and online knowledge repositories: A need-finding study.}
\newblock In \emph{Proceedings of the 2024 CHI Conference on Human Factors in Computing Systems}, CHI '24, New York, NY, USA. Association for Computing Machinery.

\bibitem[{Ogunremi et~al.(2023)Ogunremi, Jurafsky, and Manning}]{ogunremi-etal-2023-mini}
Tolulope Ogunremi, Dan Jurafsky, and Christopher Manning. 2023.
\newblock \href {https://doi.org/10.18653/v1/2023.findings-eacl.93} {Mini but mighty: Efficient multilingual pretraining with linguistically-informed data selection}.
\newblock In \emph{Findings of the Association for Computational Linguistics: EACL 2023}, pages 1251--1266, Dubrovnik, Croatia. Association for Computational Linguistics.

\bibitem[{Ojo et~al.(2024)Ojo, Ogueji, Stenetorp, and Adelani}]{ojo2024goodlargelanguagemodels}
Jessica Ojo, Kelechi Ogueji, Pontus Stenetorp, and David~Ifeoluwa Adelani. 2024.
\newblock \href {http://arxiv.org/abs/2311.07978} {How good are large language models on african languages?}

\bibitem[{Oladipo et~al.(2023)Oladipo, Adeyemi, Ahia, Owodunni, Ogundepo, Adelani, and Lin}]{oladipo2023better}
Akintunde Oladipo, Mofetoluwa Adeyemi, Orevaoghene Ahia, Abraham~Toluwase Owodunni, Odunayo Ogundepo, David~Ifeoluwa Adelani, and Jimmy Lin. 2023.
\newblock \href {https://openreview.net/forum?id=ybc9V6Cbq2} {Better quality pre-training data and t5 models for african languages}.
\newblock In \emph{The 2023 Conference on Empirical Methods in Natural Language Processing}.

\bibitem[{OpenAI et~al.(2024)OpenAI, Achiam, Adler, Agarwal, Ahmad, Akkaya, Aleman, Almeida, Altenschmidt, Altman, Anadkat, Avila, Babuschkin, Balaji, Balcom, Baltescu, Bao, Bavarian, Belgum, Bello, Berdine, Bernadett-Shapiro, Berner, Bogdonoff, Boiko, Boyd, Brakman, Brockman, Brooks, Brundage, Button, Cai, Campbell, Cann, Carey, Carlson, Carmichael, Chan, Chang, Chantzis, Chen, Chen, Chen, Chen, Chen, Chess, Cho, Chu, Chung, Cummings, Currier, Dai, Decareaux, Degry, Deutsch, Deville, Dhar, Dohan, Dowling, Dunning, Ecoffet, Eleti, Eloundou, Farhi, Fedus, Felix, Fishman, Forte, Fulford, Gao, Georges, Gibson, Goel, Gogineni, Goh, Gontijo-Lopes, Gordon, Grafstein, Gray, Greene, Gross, Gu, Guo, Hallacy, Han, Harris, He, Heaton, Heidecke, Hesse, Hickey, Hickey, Hoeschele, Houghton, Hsu, Hu, Hu, Huizinga, Jain, Jain, Jang, Jiang, Jiang, Jin, Jin, Jomoto, Jonn, Jun, Kaftan, Łukasz Kaiser, Kamali, Kanitscheider, Keskar, Khan, Kilpatrick, Kim, Kim, Kim, Kirchner, Kiros, Knight, Kokotajlo, Łukasz Kondraciuk,
  Kondrich, Konstantinidis, Kosic, Krueger, Kuo, Lampe, Lan, Lee, Leike, Leung, Levy, Li, Lim, Lin, Lin, Litwin, Lopez, Lowe, Lue, Makanju, Malfacini, Manning, Markov, Markovski, Martin, Mayer, Mayne, McGrew, McKinney, McLeavey, McMillan, McNeil, Medina, Mehta, Menick, Metz, Mishchenko, Mishkin, Monaco, Morikawa, Mossing, Mu, Murati, Murk, Mély, Nair, Nakano, Nayak, Neelakantan, Ngo, Noh, Ouyang, O'Keefe, Pachocki, Paino, Palermo, Pantuliano, Parascandolo, Parish, Parparita, Passos, Pavlov, Peng, Perelman, de~Avila Belbute~Peres, Petrov, de~Oliveira~Pinto, Michael, Pokorny, Pokrass, Pong, Powell, Power, Power, Proehl, Puri, Radford, Rae, Ramesh, Raymond, Real, Rimbach, Ross, Rotsted, Roussez, Ryder, Saltarelli, Sanders, Santurkar, Sastry, Schmidt, Schnurr, Schulman, Selsam, Sheppard, Sherbakov, Shieh, Shoker, Shyam, Sidor, Sigler, Simens, Sitkin, Slama, Sohl, Sokolowsky, Song, Staudacher, Such, Summers, Sutskever, Tang, Tezak, Thompson, Tillet, Tootoonchian, Tseng, Tuggle, Turley, Tworek, Uribe, Vallone,
  Vijayvergiya, Voss, Wainwright, Wang, Wang, Wang, Ward, Wei, Weinmann, Welihinda, Welinder, Weng, Weng, Wiethoff, Willner, Winter, Wolrich, Wong, Workman, Wu, Wu, Wu, Xiao, Xu, Yoo, Yu, Yuan, Zaremba, Zellers, Zhang, Zhang, Zhao, Zheng, Zhuang, Zhuk, and Zoph}]{openai2024gpt4technicalreport}
OpenAI, Josh Achiam, Steven Adler, Sandhini Agarwal, Lama Ahmad, Ilge Akkaya, Florencia~Leoni Aleman, Diogo Almeida, Janko Altenschmidt, Sam Altman, Shyamal Anadkat, Red Avila, Igor Babuschkin, Suchir Balaji, Valerie Balcom, Paul Baltescu, Haiming Bao, Mohammad Bavarian, Jeff Belgum, Irwan Bello, Jake Berdine, Gabriel Bernadett-Shapiro, Christopher Berner, Lenny Bogdonoff, Oleg Boiko, Madelaine Boyd, Anna-Luisa Brakman, Greg Brockman, Tim Brooks, Miles Brundage, Kevin Button, Trevor Cai, Rosie Campbell, Andrew Cann, Brittany Carey, Chelsea Carlson, Rory Carmichael, Brooke Chan, Che Chang, Fotis Chantzis, Derek Chen, Sully Chen, Ruby Chen, Jason Chen, Mark Chen, Ben Chess, Chester Cho, Casey Chu, Hyung~Won Chung, Dave Cummings, Jeremiah Currier, Yunxing Dai, Cory Decareaux, Thomas Degry, Noah Deutsch, Damien Deville, Arka Dhar, David Dohan, Steve Dowling, Sheila Dunning, Adrien Ecoffet, Atty Eleti, Tyna Eloundou, David Farhi, Liam Fedus, Niko Felix, Simón~Posada Fishman, Juston Forte, Isabella Fulford, Leo
  Gao, Elie Georges, Christian Gibson, Vik Goel, Tarun Gogineni, Gabriel Goh, Rapha Gontijo-Lopes, Jonathan Gordon, Morgan Grafstein, Scott Gray, Ryan Greene, Joshua Gross, Shixiang~Shane Gu, Yufei Guo, Chris Hallacy, Jesse Han, Jeff Harris, Yuchen He, Mike Heaton, Johannes Heidecke, Chris Hesse, Alan Hickey, Wade Hickey, Peter Hoeschele, Brandon Houghton, Kenny Hsu, Shengli Hu, Xin Hu, Joost Huizinga, Shantanu Jain, Shawn Jain, Joanne Jang, Angela Jiang, Roger Jiang, Haozhun Jin, Denny Jin, Shino Jomoto, Billie Jonn, Heewoo Jun, Tomer Kaftan, Łukasz Kaiser, Ali Kamali, Ingmar Kanitscheider, Nitish~Shirish Keskar, Tabarak Khan, Logan Kilpatrick, Jong~Wook Kim, Christina Kim, Yongjik Kim, Jan~Hendrik Kirchner, Jamie Kiros, Matt Knight, Daniel Kokotajlo, Łukasz Kondraciuk, Andrew Kondrich, Aris Konstantinidis, Kyle Kosic, Gretchen Krueger, Vishal Kuo, Michael Lampe, Ikai Lan, Teddy Lee, Jan Leike, Jade Leung, Daniel Levy, Chak~Ming Li, Rachel Lim, Molly Lin, Stephanie Lin, Mateusz Litwin, Theresa Lopez, Ryan
  Lowe, Patricia Lue, Anna Makanju, Kim Malfacini, Sam Manning, Todor Markov, Yaniv Markovski, Bianca Martin, Katie Mayer, Andrew Mayne, Bob McGrew, Scott~Mayer McKinney, Christine McLeavey, Paul McMillan, Jake McNeil, David Medina, Aalok Mehta, Jacob Menick, Luke Metz, Andrey Mishchenko, Pamela Mishkin, Vinnie Monaco, Evan Morikawa, Daniel Mossing, Tong Mu, Mira Murati, Oleg Murk, David Mély, Ashvin Nair, Reiichiro Nakano, Rajeev Nayak, Arvind Neelakantan, Richard Ngo, Hyeonwoo Noh, Long Ouyang, Cullen O'Keefe, Jakub Pachocki, Alex Paino, Joe Palermo, Ashley Pantuliano, Giambattista Parascandolo, Joel Parish, Emy Parparita, Alex Passos, Mikhail Pavlov, Andrew Peng, Adam Perelman, Filipe de~Avila Belbute~Peres, Michael Petrov, Henrique~Ponde de~Oliveira~Pinto, Michael, Pokorny, Michelle Pokrass, Vitchyr~H. Pong, Tolly Powell, Alethea Power, Boris Power, Elizabeth Proehl, Raul Puri, Alec Radford, Jack Rae, Aditya Ramesh, Cameron Raymond, Francis Real, Kendra Rimbach, Carl Ross, Bob Rotsted, Henri Roussez,
  Nick Ryder, Mario Saltarelli, Ted Sanders, Shibani Santurkar, Girish Sastry, Heather Schmidt, David Schnurr, John Schulman, Daniel Selsam, Kyla Sheppard, Toki Sherbakov, Jessica Shieh, Sarah Shoker, Pranav Shyam, Szymon Sidor, Eric Sigler, Maddie Simens, Jordan Sitkin, Katarina Slama, Ian Sohl, Benjamin Sokolowsky, Yang Song, Natalie Staudacher, Felipe~Petroski Such, Natalie Summers, Ilya Sutskever, Jie Tang, Nikolas Tezak, Madeleine~B. Thompson, Phil Tillet, Amin Tootoonchian, Elizabeth Tseng, Preston Tuggle, Nick Turley, Jerry Tworek, Juan Felipe~Cerón Uribe, Andrea Vallone, Arun Vijayvergiya, Chelsea Voss, Carroll Wainwright, Justin~Jay Wang, Alvin Wang, Ben Wang, Jonathan Ward, Jason Wei, CJ~Weinmann, Akila Welihinda, Peter Welinder, Jiayi Weng, Lilian Weng, Matt Wiethoff, Dave Willner, Clemens Winter, Samuel Wolrich, Hannah Wong, Lauren Workman, Sherwin Wu, Jeff Wu, Michael Wu, Kai Xiao, Tao Xu, Sarah Yoo, Kevin Yu, Qiming Yuan, Wojciech Zaremba, Rowan Zellers, Chong Zhang, Marvin Zhang, Shengjia
  Zhao, Tianhao Zheng, Juntang Zhuang, William Zhuk, and Barret Zoph. 2024.
\newblock \href {http://arxiv.org/abs/2303.08774} {Gpt-4 technical report}.

\bibitem[{Pan et~al.(2024)Pan, Luo, Wang, Chen, Wang, and Wu}]{llm_kg}
Shirui Pan, Linhao Luo, Yufei Wang, Chen Chen, Jiapu Wang, and Xindong Wu. 2024.
\newblock Unifying large language models and knowledge graphs: A roadmap.
\newblock \emph{IEEE Transactions on Knowledge and Data Engineering (TKDE)}.

\bibitem[{Paulheim(2018)}]{paulheim_2018}
Heiko Paulheim. 2018.
\newblock \href {https://ceur-ws.org/Vol-2180/ISWC_2018_Outrageous_Ideas_paper_10.pdf} {How much is a triple? estimating the cost of knowledge graph creation}.

\bibitem[{Petroni et~al.(2019)Petroni, Rockt{\"a}schel, Riedel, Lewis, Bakhtin, Wu, and Miller}]{petroni-etal-2019-language}
Fabio Petroni, Tim Rockt{\"a}schel, Sebastian Riedel, Patrick Lewis, Anton Bakhtin, Yuxiang Wu, and Alexander Miller. 2019.
\newblock \href {https://doi.org/10.18653/v1/D19-1250} {Language models as knowledge bases?}
\newblock In \emph{Proceedings of the 2019 Conference on Empirical Methods in Natural Language Processing and the 9th International Joint Conference on Natural Language Processing (EMNLP-IJCNLP)}, pages 2463--2473, Hong Kong, China. Association for Computational Linguistics.

\bibitem[{Reimers and Gurevych(2019)}]{reimers-2019-sentence-bert}
Nils Reimers and Iryna Gurevych. 2019.
\newblock \href {https://arxiv.org/abs/1908.10084} {Sentence-bert: Sentence embeddings using siamese bert-networks}.
\newblock In \emph{Proceedings of the 2019 Conference on Empirical Methods in Natural Language Processing}. Association for Computational Linguistics.

\bibitem[{Reinanda et~al.(2020)Reinanda, Meij, and de~Rijke}]{kg-ir-2020-Reinanda}
Ridho Reinanda, Edgar Meij, and Maarten de~Rijke. 2020.
\newblock \href {https://doi.org/10.1561/1500000063} {Knowledge graphs: An information retrieval perspective}.
\newblock \emph{Found. Trends Inf. Retr.}, 14(4):289–444.

\bibitem[{Robertson and Zaragoza(2009)}]{INR-019}
Stephen Robertson and Hugo Zaragoza. 2009.
\newblock \href {https://doi.org/10.1561/1500000019} {The probabilistic relevance framework: Bm25 and beyond}.
\newblock \emph{Foundations and Trends® in Information Retrieval}, 3(4):333--389.

\bibitem[{Samir et~al.(2024)Samir, Park, Field, Shwartz, and Tsvetkov}]{samir-etal-2024-locating}
Farhan Samir, Chan~Young Park, Anjalie Field, Vered Shwartz, and Yulia Tsvetkov. 2024.
\newblock \href {https://doi.org/10.18653/v1/2024.emnlp-main.384} {Locating information gaps and narrative inconsistencies across languages: A case study of {LGBT} people portrayals on {W}ikipedia}.
\newblock In \emph{Proceedings of the 2024 Conference on Empirical Methods in Natural Language Processing}, pages 6747--6762, Miami, Florida, USA. Association for Computational Linguistics.

\bibitem[{Saxena et~al.(2022{\natexlab{a}})Saxena, Kochsiek, and Gemulla}]{kgt5-saxena2022sequence}
Apoorv Saxena, Adrian Kochsiek, and Rainer Gemulla. 2022{\natexlab{a}}.
\newblock Sequence-to-sequence knowledge graph completion and question answering.
\newblock In \emph{Proceedings of the 60th Annual Meeting of the Association for Computational Linguistics (Volume 1: Long Papers)}, pages 2814--2828.

\bibitem[{Saxena et~al.(2022{\natexlab{b}})Saxena, Kochsiek, and Gemulla}]{saxena-etal-2022-sequence}
Apoorv Saxena, Adrian Kochsiek, and Rainer Gemulla. 2022{\natexlab{b}}.
\newblock \href {https://doi.org/10.18653/v1/2022.acl-long.201} {Sequence-to-sequence knowledge graph completion and question answering}.
\newblock In \emph{Proceedings of the 60th Annual Meeting of the Association for Computational Linguistics (Volume 1: Long Papers)}, pages 2814--2828, Dublin, Ireland. Association for Computational Linguistics.

\bibitem[{Song et~al.(2023)Song, He, Gao, Cai, Liu, Zhengtao, and Zhao}]{song2023multilingual}
Ran Song, Shizhu He, Shengxiang Gao, Li~Cai, Kang Liu, YU~Zhengtao, and Jun Zhao. 2023.
\newblock Multilingual knowledge graph completion from pretrained language models with knowledge constraints.
\newblock In \emph{The 61st Annual Meeting Of The Association For Computational Linguistics}.

\bibitem[{Sun et~al.(2020)Sun, Wang, Hu, Chen, Dai, Zhang, and Qu}]{sun2020knowledge}
Zequn Sun, Chengming Wang, Wei Hu, Muhao Chen, Jian Dai, Wei Zhang, and Yuzhong Qu. 2020.
\newblock Knowledge graph alignment network with gated multi-hop neighborhood aggregation.
\newblock In \emph{Proceedings of the AAAI conference on artificial intelligence}, volume~34, pages 222--229.

\bibitem[{Tian et~al.(2024)Tian, Luo, Xu, Yuan, Jiang, Wei, and Wang}]{tian2024kg}
Shiyu Tian, Yangyang Luo, Tianze Xu, Caixia Yuan, Huixing Jiang, Chen Wei, and Xiaojie Wang. 2024.
\newblock Kg-adapter: Enabling knowledge graph integration in large language models through parameter-efficient fine-tuning.
\newblock In \emph{Findings of the Association for Computational Linguistics ACL 2024}, pages 3813--3828.

\bibitem[{{\"U}st{\"u}n et~al.(2024){\"U}st{\"u}n, Aryabumi, Yong, Ko, D{'}souza, Onilude, Bhandari, Singh, Ooi, Kayid, Vargus, Blunsom, Longpre, Muennighoff, Fadaee, Kreutzer, and Hooker}]{ustun-etal-2024-aya}
Ahmet {\"U}st{\"u}n, Viraat Aryabumi, Zheng Yong, Wei-Yin Ko, Daniel D{'}souza, Gbemileke Onilude, Neel Bhandari, Shivalika Singh, Hui-Lee Ooi, Amr Kayid, Freddie Vargus, Phil Blunsom, Shayne Longpre, Niklas Muennighoff, Marzieh Fadaee, Julia Kreutzer, and Sara Hooker. 2024.
\newblock \href {https://doi.org/10.18653/v1/2024.acl-long.845} {Aya model: An instruction finetuned open-access multilingual language model}.
\newblock In \emph{Proceedings of the 62nd Annual Meeting of the Association for Computational Linguistics (Volume 1: Long Papers)}, pages 15894--15939, Bangkok, Thailand. Association for Computational Linguistics.

\bibitem[{Vrande{\v{c}}i{\'c} and Kr{\"o}tzsch(2014)}]{vrandevcic2014wikidata}
Denny Vrande{\v{c}}i{\'c} and Markus Kr{\"o}tzsch. 2014.
\newblock Wikidata: a free collaborative knowledgebase.
\newblock \emph{Communications of the ACM}, 57(10):78--85.

\bibitem[{Vrande\v{c}i\'{c} and Kr\"{o}tzsch(2014)}]{10.1145/2629489}
Denny Vrande\v{c}i\'{c} and Markus Kr\"{o}tzsch. 2014.
\newblock \href {https://doi.org/10.1145/2629489} {Wikidata: a free collaborative knowledgebase}.
\newblock \emph{Commun. ACM}, 57(10):78–85.

\bibitem[{Wang et~al.(2021)Wang, Gao, Zhu, Zhang, Liu, Li, and Tang}]{wang-etal-2021-kepler}
Xiaozhi Wang, Tianyu Gao, Zhaocheng Zhu, Zhengyan Zhang, Zhiyuan Liu, Juanzi Li, and Jian Tang. 2021.
\newblock \href {https://doi.org/10.1162/tacl_a_00360} {{KEPLER}: A unified model for knowledge embedding and pre-trained language representation}.
\newblock \emph{Transactions of the Association for Computational Linguistics}, 9:176--194.

\bibitem[{Wu et~al.(2022)Wu, Zhao, Hu, Minervini, Stenetorp, and Riedel}]{wu2022efficient}
Yuxiang Wu, Yu~Zhao, Baotian Hu, Pasquale Minervini, Pontus Stenetorp, and Sebastian Riedel. 2022.
\newblock An efficient memory-augmented transformer for knowledge-intensive nlp tasks.
\newblock In \emph{Proceedings of the 2022 Conference on Empirical Methods in Natural Language Processing}, pages 5184--5196.

\bibitem[{Xue et~al.(2021)Xue, Constant, Roberts, Kale, Al-Rfou, Siddhant, Barua, and Raffel}]{xue-etal-2021-mt5}
Linting Xue, Noah Constant, Adam Roberts, Mihir Kale, Rami Al-Rfou, Aditya Siddhant, Aditya Barua, and Colin Raffel. 2021.
\newblock \href {https://doi.org/10.18653/v1/2021.naacl-main.41} {m{T}5: A massively multilingual pre-trained text-to-text transformer}.
\newblock In \emph{Proceedings of the 2021 Conference of the North American Chapter of the Association for Computational Linguistics: Human Language Technologies}, pages 483--498, Online. Association for Computational Linguistics.

\bibitem[{Yang et~al.(2024)Yang, Chen, Li, Ding, and Wu}]{yang2024give}
Linyao Yang, Hongyang Chen, Zhao Li, Xiao Ding, and Xindong Wu. 2024.
\newblock Give us the facts: Enhancing large language models with knowledge graphs for fact-aware language modeling.
\newblock \emph{IEEE Transactions on Knowledge and Data Engineering}.

\bibitem[{Yao et~al.(2019)Yao, Mao, and Luo}]{yao2019kgbertbertknowledgegraph}
Liang Yao, Chengsheng Mao, and Yuan Luo. 2019.
\newblock \href {http://arxiv.org/abs/1909.03193} {Kg-bert: Bert for knowledge graph completion}.

\bibitem[{Yimam et~al.(2020)Yimam, Alemayehu, Ayele, and Biemann}]{yimam-etal-2020-exploring}
Seid~Muhie Yimam, Hizkiel~Mitiku Alemayehu, Abinew Ayele, and Chris Biemann. 2020.
\newblock \href {https://doi.org/10.18653/v1/2020.coling-main.91} {Exploring {A}mharic sentiment analysis from social media texts: Building annotation tools and classification models}.
\newblock In \emph{Proceedings of the 28th International Conference on Computational Linguistics}, pages 1048--1060, Barcelona, Spain (Online). International Committee on Computational Linguistics.

\bibitem[{Yu et~al.(2024)Yu, Atanasova, and Augenstein}]{yu-etal-2024-revealing}
Haeun Yu, Pepa Atanasova, and Isabelle Augenstein. 2024.
\newblock \href {https://doi.org/10.18653/v1/2024.acl-long.444} {Revealing the parametric knowledge of language models: A unified framework for attribution methods}.
\newblock In \emph{Proceedings of the 62nd Annual Meeting of the Association for Computational Linguistics (Volume 1: Long Papers)}, pages 8173--8186, Bangkok, Thailand. Association for Computational Linguistics.

\bibitem[{Zhong et~al.(2023)Zhong, Wu, Li, Peng, and Wu}]{zhong2023comprehensive}
Lingfeng Zhong, Jia Wu, Qian Li, Hao Peng, and Xindong Wu. 2023.
\newblock A comprehensive survey on automatic knowledge graph construction.
\newblock \emph{ACM Computing Surveys}, 56(4):1--62.

\bibitem[{Zhou et~al.(2022)Zhou, Liu, Vuli{\'c}, Collier, and Chen}]{zhou-etal-2022-prix}
Wenxuan Zhou, Fangyu Liu, Ivan Vuli{\'c}, Nigel Collier, and Muhao Chen. 2022.
\newblock \href {https://doi.org/10.18653/v1/2022.acl-long.371} {Prix-{LM}: Pretraining for multilingual knowledge base construction}.
\newblock In \emph{Proceedings of the 60th Annual Meeting of the Association for Computational Linguistics (Volume 1: Long Papers)}, pages 5412--5424, Dublin, Ireland. Association for Computational Linguistics.

\end{thebibliography}
\bibliographystyle{acl_natbib}

\appendix \label{sec:appendix}

\section{More Results} \label{apn:more_results}
\subsection{(im)perfect Retriever}
Since we did not have labeled data, we used the (im)perfect retriever as an upper bound for performance. This allowed us to interrogate how well our system performs in an ideal case and to perform a set of ablation studies. To check how well the (im)perfect retriever performs compared to the other retrievers in our experiments, we compare the performance of the (im)perfect retriever with the LaBSE retriever breaking down the results by context language. As Table \ref{tab:upperbound} shows, we find that the (im)perfect retriever outperforms the LaBSE retriever regardless of what language the context is in. However, the (im)perfect retriever, which searches for the tail entity in the Wikipedia article of the head entity, may not always retrieve context (e.g if the head entity Wiki article does not explicitly mention the tail entity). In the cases where the (im)perfect retriever does not fetch context, the LaBSE retriever outperforms. Nonetheless, the (im)perfect retriever provides an upper bound for the cases where context is provided. 
\begin{table}[]
    \centering
    \begin{tabular}{p{2.3cm}|p{2cm}|c}
    \toprule
        Context Language & (im)perfect retriever & LaBSE \\
    \midrule
        Tir & 78.05 & 43.90 \\
        Amh & 64.19 & 19.75 \\
        Ara & 58.87 & 16.45 \\
        Eng & 64.51 & 22.58 \\
        \midrule
        No-context in (im)perfect & 9.09 & 12.53 \\
        \bottomrule
    \end{tabular}
    \caption{Performance comparison of the (im)perfect retriever and the LaBSE retriever broken down by context language.}
    \label{tab:upperbound}
\end{table}
\subsection{Multilingual Context}
\begin{figure}
    \centering
    \subfloat{%
        \includegraphics[width=0.25\textwidth]{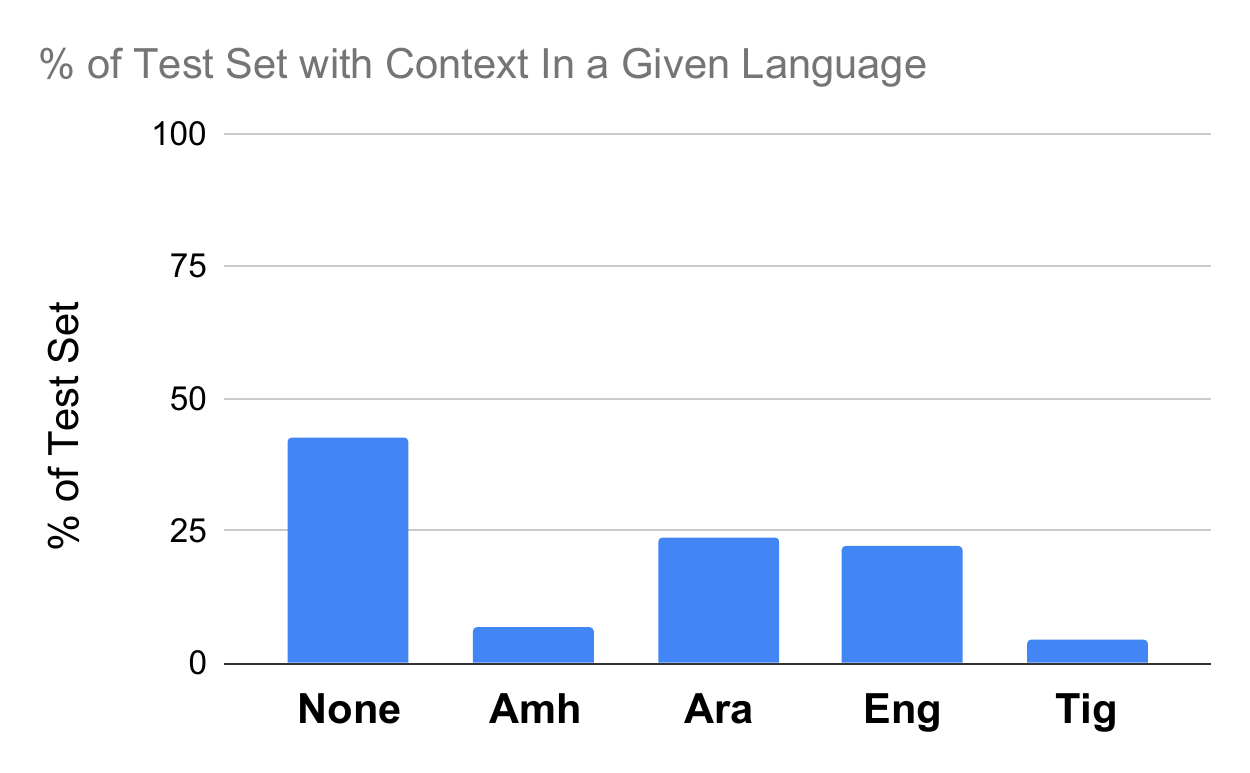}
        \label{fig:a}}
    \subfloat{%
        \includegraphics[width=0.25\textwidth]{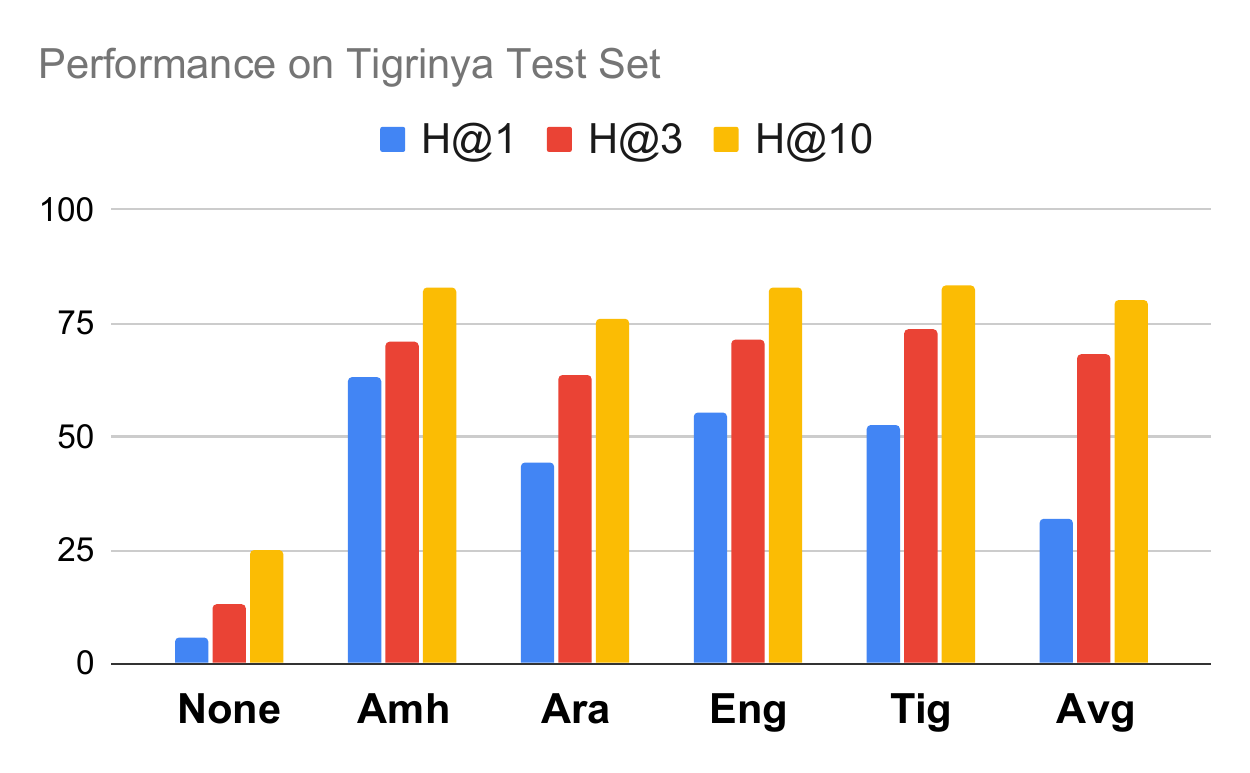}
        \label{fig:b}}
    \caption{Effects of context from different languages on Hits for Tigrinya.}
    \label{fig:tigrinya_breakdown}
\end{figure}

\begin{figure}
    \centering
    \subfloat{%
        \includegraphics[width=0.25\textwidth]{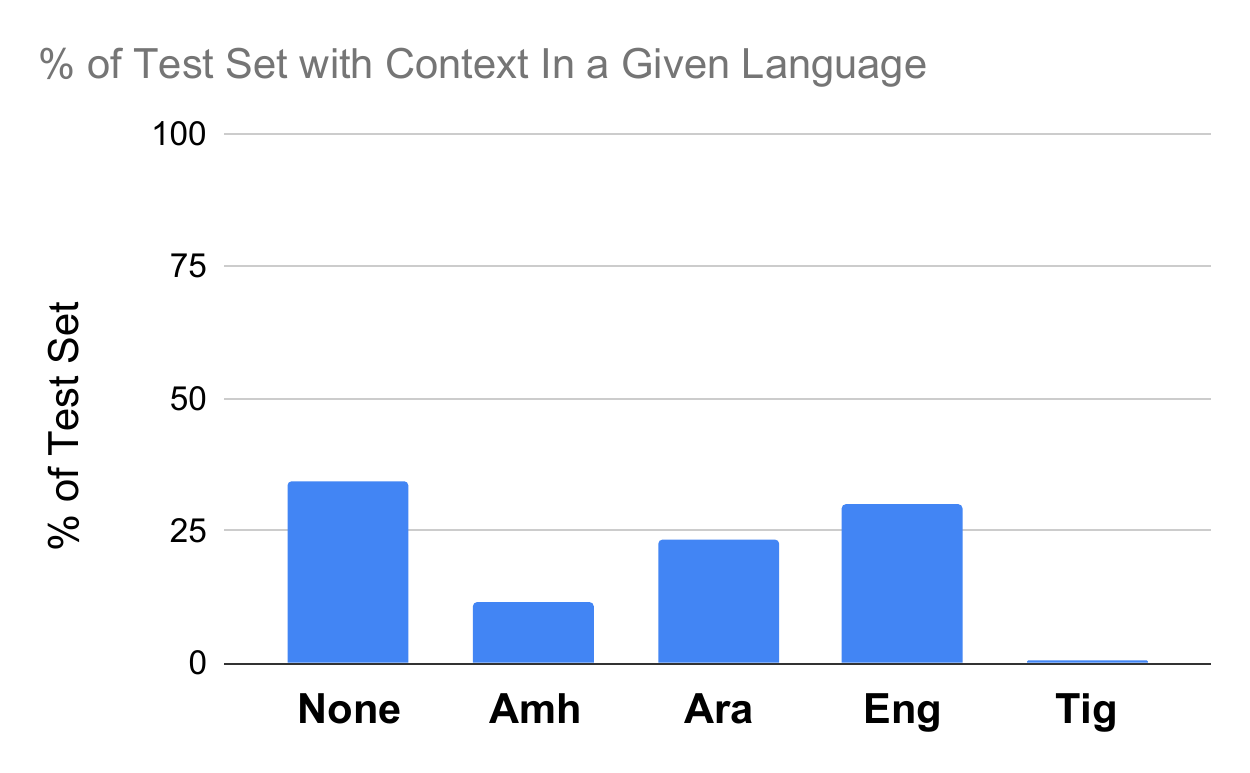}
        \label{fig:a}}
    \subfloat{%
        \includegraphics[width=0.25\textwidth]{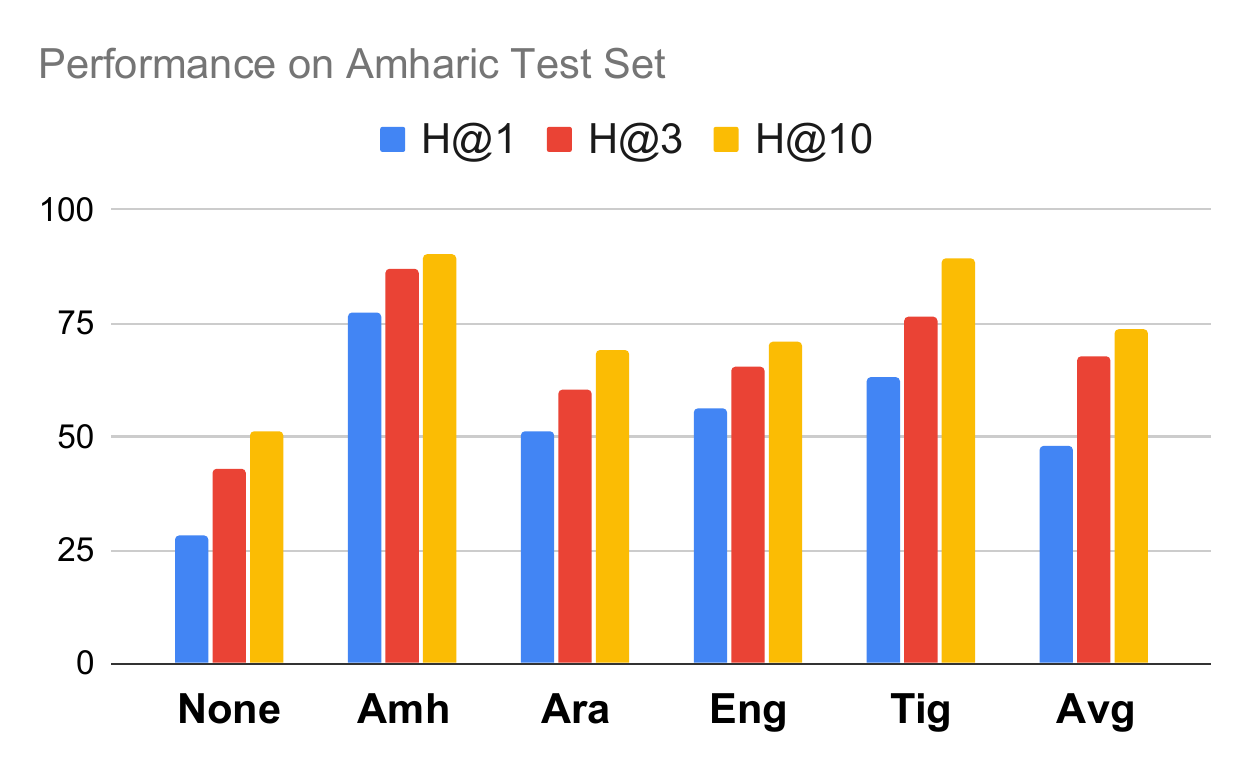}
        \label{fig:b}}
    \caption{Effects of context from different languages on Hits for Amharic.}
    \label{fig:amharic_breakdown}
\end{figure}

In this section, we provide figures and graphs to support the results reported in Sec. \label{sec:additional}. Figure \ref{fig:tigrinya_breakdown} and Figure \ref{fig:amharic_breakdown} give a breakdown of performance by context language along with what percentage of the context comes from which language. 

\subsection{Qualitative Examples}
\begin{figure}
    \centering
    \includegraphics[width=\linewidth]{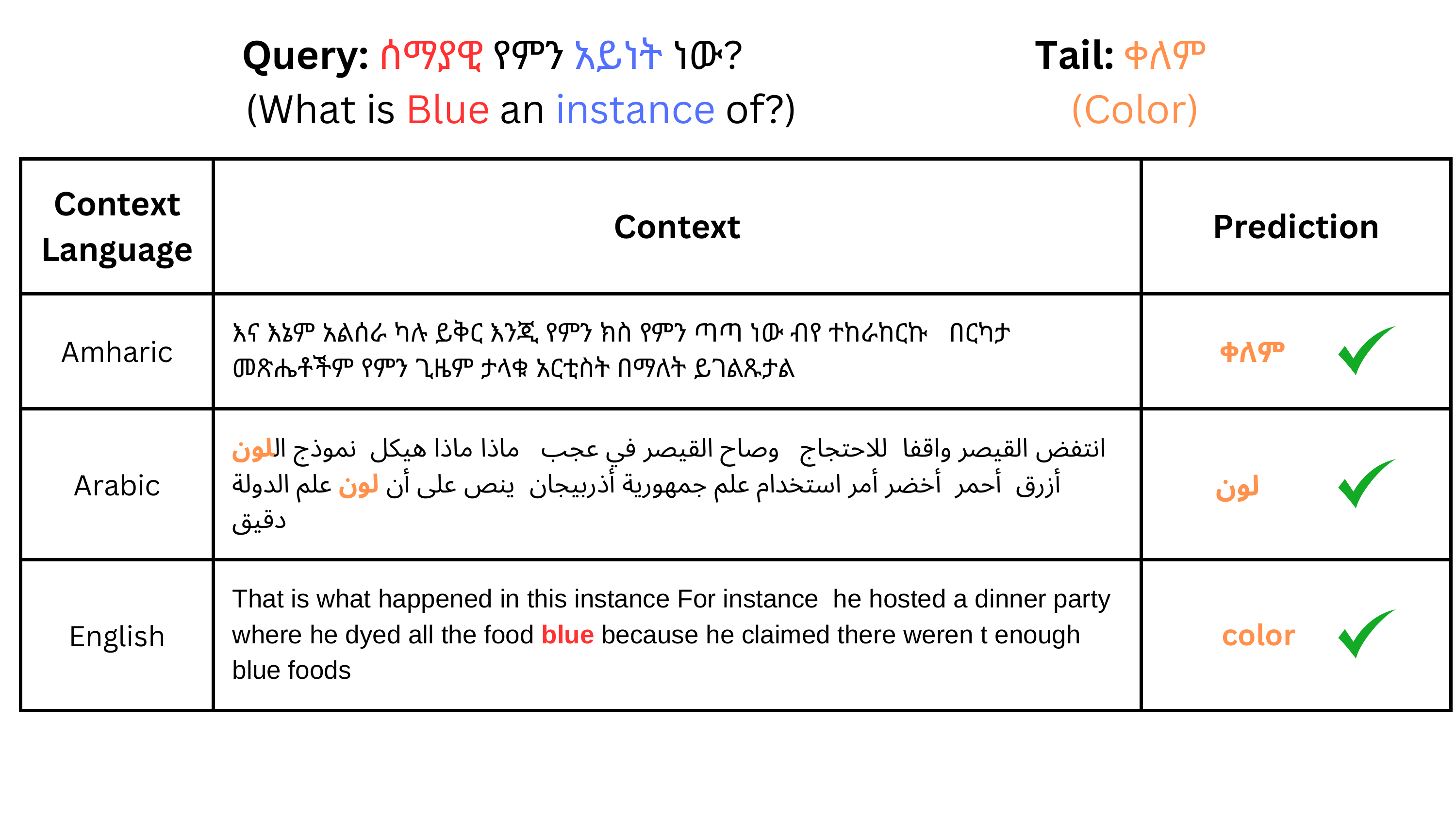}
    \caption{Context from Arabic contains the tail entity but context from Amharic and English does not; however, as a result of cross-lingual alignment, the generative model is correctly able to predict the tail entity in all languages.}
    \label{fig:multilingual}
\end{figure}

In this section, we provide qualitative evidence that explains our system performance. In Figure \ref{fig:multilingual}, we show an example that demonstrates how our system does cross-lingual entity alignment. Only the Arabic context includes the target tail entity. However, the generative model is able to predict the tail entity in all three languages, suggesting that it was able to correctly align the entity `color' in all the languages. 

\begin{figure}
    \centering
    \includegraphics[width=\linewidth]{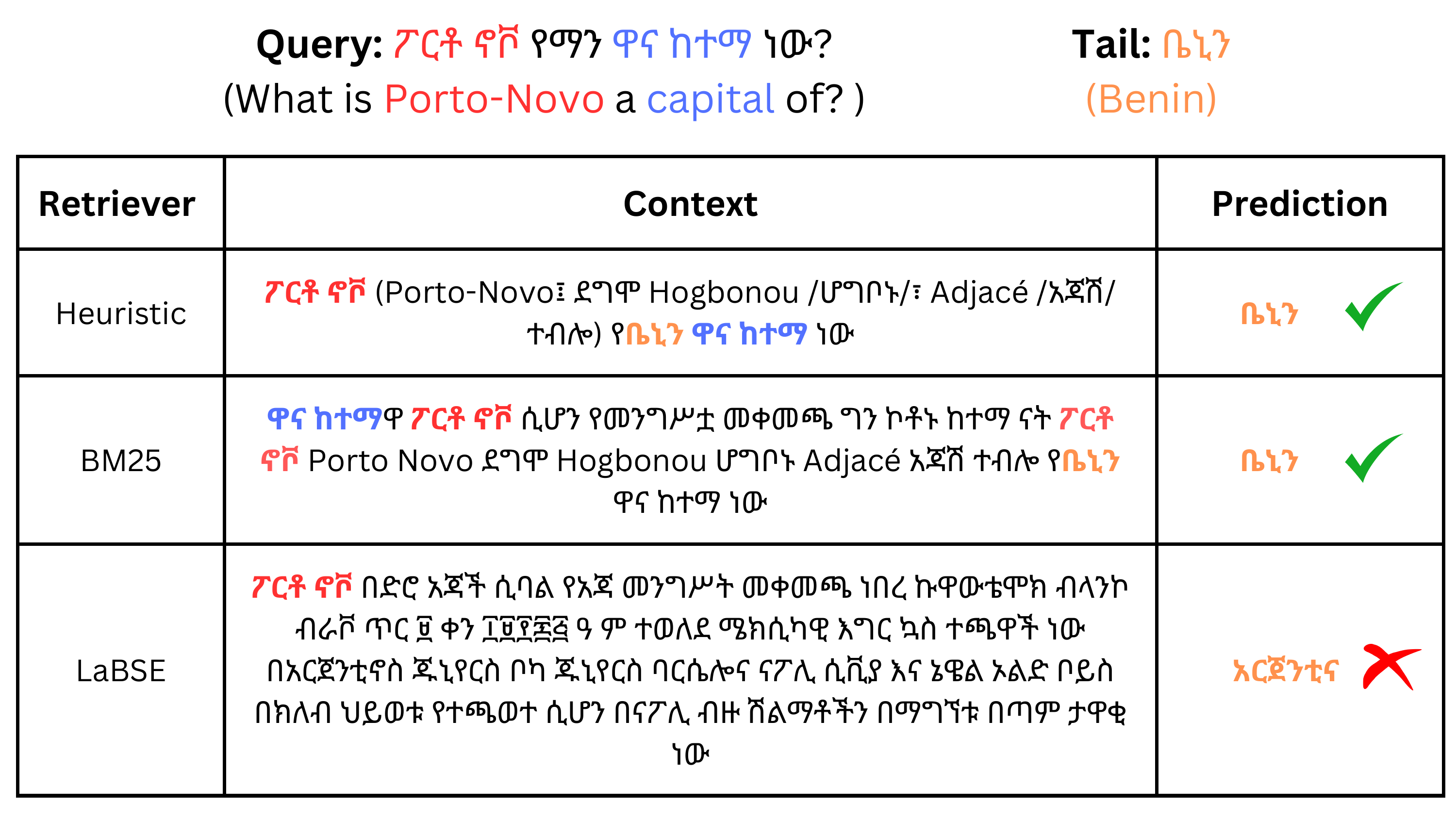}
    \caption{Example where the BM25 and Heuristic retriever were able to fetch context that includes the head, relation, and tail while the LaBSE model context only contains the head entity. As a result, the generative model makes an incorrect prediction.}
    \label{fig:different_performance}
\end{figure}

In Figure \ref{fig:different_performance}, we observe an example where the Heuristic (im)perfect retriever and the BM25 retrieved the correct context, while the LaBSE retriever did not. Both the (im)perfect retriever and BM25 were able to find a context that included the head, relation, and tail; while the LaBSE model context only has the head entity. The first sentence retrieved by the LaBSE model translates to ``Porto-Novo, which was known as Ajashe, was the main location for the Aja government.'' While the context contains information about the head entity, it does not answer the query ``What is Porto-Novo a capital of?'' In this case, the prediction of the generative model with the LaBSE model was incorrect. However, since our approach is modular, the retriever can be improved separately which will increase performance. 

\section{Further Details on Methods} \label{apn:more_method_details}

Here, we provide details on how we trained our \textit{generator} and \textit{retriever} models. For BM25, we used the library by \citet{lù2024bm25sordersmagnitudefaster} with default settings. Below, we describe our finetuning setup for LaBSE and AfriTeva.
\subsection{Finetuning LaBSE} \label{apn:finetune_labse}
While the LaBSE model~\cite{feng-etal-2022-language} includes Amharic, Arabic and English, it does not include Tigrinya. Since we do not have labeled data for finetuning, we used the contexts from our (im)perfect retriever to prepare training data for the LaBSE model. The (im)perfect retriever extracts sentences that have the tail entity from the introduction paragraph of the Wikipedia article for the head entity. We call this retriever (im)perfect because:
\begin{itemize}
    \item It may not always retrieve context. For instance, if the tail entity is not mentioned at all in the introduction paragraph of the Wikipedia article, the (im)perfect retriever will not return anything.
    \item Some of the sentences it retrieves may not be a direct answer to the query. For example, if the question is ``What is \textcolor{red}{Surafel Dagnachew}'s \textcolor{blue}{place of birth}?'', the retrieved sentence may be ``\textcolor{red}{Surafel Dagnachew} plays for the football team of \textcolor{orange}{Ethiopia}.''. While it does not directly answer the question, it does include the tail entity. 
\end{itemize}

Using the training data of both Amharic and Tigrinya KG with context retrieved by the (im)perfect retriever, we finetune the LaBSE model. We use contrastive loss during training. Training the model requires an \textit{anchor}, which is the query we are using for retrieval, and \textit{positive} and \textit{negative} examples. The extracted context from the (im)perfect retriever serve as the positive example; i.e the model learns to increase the similarity score between the anchor and this positive example. To walk us through a training step, let us take the triple (\textcolor{red}{Surafel Dagnachew}, \textcolor{blue}{place of birth}, \textcolor{orange}{Ethiopia}). Once reformulated to question answering format, teh training data point becomes <What is \textcolor{red}{Surafel Dagnachew}'s \textcolor{blue}{place of birth}?, \textcolor{orange}{Ethiopia}>. For the \textit{negative} examples, we first get all the sentences that do not include the tail entity (in this case, \textcolor{orange}{Ethiopia}). We then prepare three types of negative examples: 
\begin{itemize}
    \item Hard Negative: A sentence that does not have any of the entities or relation (head and tail entities or relation). (e.g ``Barack Obama was the president of the United States of America.'')
    \item Head Negative: A sentence that contains neither the tail entity nor the head entity but contains the relation. (e.g ``Barack Obama's \textcolor{blue}{place of birth} is the United States of America.'')
    \item Relation Negative: A sentence that contains neither the tail entity nor the relation but contains the head entity. (e.g ``\textcolor{red}{Surafel Dagnachew} joined Fasil Kenema in 2018.'')

\end{itemize}

We finetuned the model for 50 epochs with a learning rate of 3e-5 and warm up for the first 15\% of the training steps. We accessed the model and conducted our finetuning through SentenceTransformers\cite{reimers-2019-sentence-bert}.
  \begin{figure*}
    \centering
    \includegraphics[width=\linewidth]{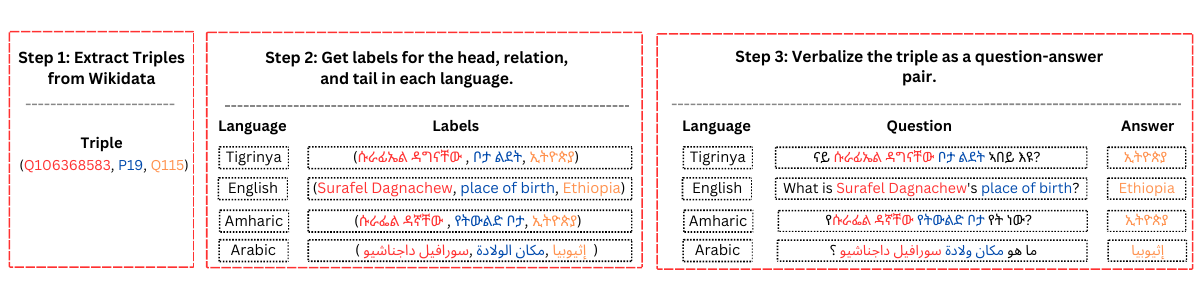}
    \caption{\textbf{Reformulating triples into question-answer pairs.} In each step depicted above, we highlight the head entity in \textcolor{red}{red}, the relation in \textcolor{blue}{blue}, and the tail entity in \textcolor{orange}{orange}. In Step 1, we start with a triple that has entity IDs and Property ID extracted from Wikidata. In Step 2, we get the labels for each of the entity and property IDs in the four languages; this gives us the textual representation of the entities and relations in the different languages. In Step 3, we plug in the head entity into the corresponding template question that has the relation; i.e the head entity \textit{Surafel Dagnachew} is plugged into the template \textit{What is \_\_\_'s place of birth?} and the tail entity, \textit{Ethiopia} is the answer the model will learn to predict. }
    \label{fig:reformulation}
\end{figure*}

\subsection {Model Training Hyperparameters}
In Table \ref{tab:params}, we give the hyperparameter details for training mT5 and AfriTeVa models. For both models, we used the base model version which has 580M and 428M parameters respectively. Training was done on two Titan RTX GPUs.
\begin{table}[]
    \centering
    \begin{tabular}{|c|c|}
    \hline
        \textbf{Parameter} & \textbf{Value} \\

    \hline
    \texttt{training\_batch\_size}    & 16 \\
    \texttt{eval\_batch\_size}    &  4 \\
    \texttt{epochs}    & 15 or 30 \\
    \texttt{learning\_rate}    & 3e-4 \\
    \texttt{lora\_rank}    &  4 \\
    \texttt{lora\_dropout}    & 0.01 \\
    
    \texttt{lora\_alpha}    & 32 \\
    \hline
    
    \end{tabular}
    \caption{Hyperparamters for training mT5 and AfriTeVa models.}
    \label{tab:params}
\end{table}

\section{Data Description} \label{apn:data}
\subsection{Manual Data Preparation}
For the Amharic data, the first author (L1 Amharic speaker) prepared the template questions manually. For the Tigrinya and English data, the first author (L2 English and L3 Tigrinya speaker) prepared the template questions. For the Tigrinya questions, two L1 speakers checked and corrected any errors. For the Arabic data, two L1 speakers created template questions. For Amharic, Tigrinya, and Arabic, we create questions in both male and female gender when necessary as the three languages are gendered. 

\subsection{Details on Knowledge Graph}

\begin{figure*}
    \centering
  \subfloat{%
        \includegraphics[width=0.5\textwidth]{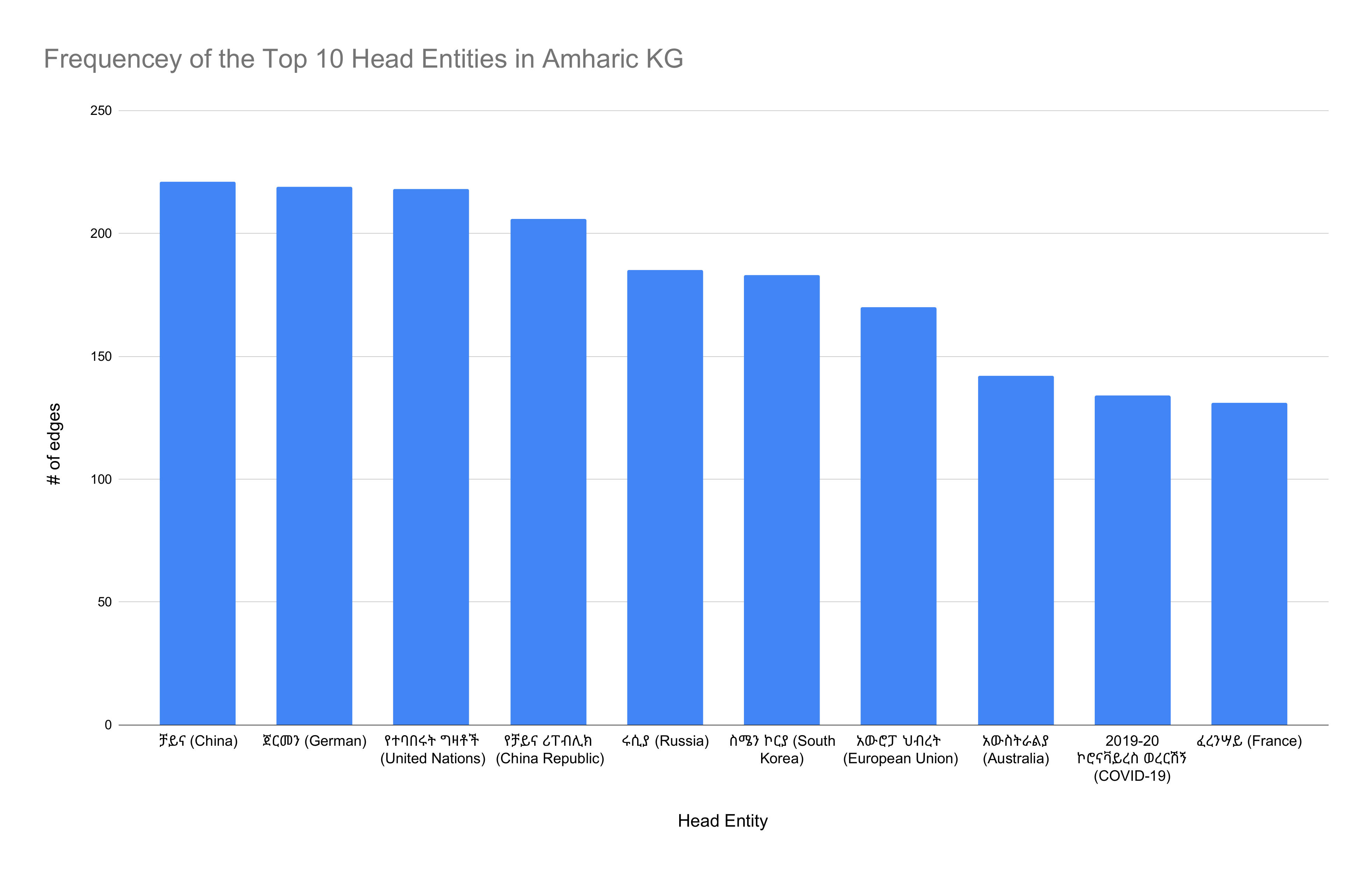}
        \label{fig:a}}
    \subfloat{%
        \includegraphics[width=0.5\textwidth]{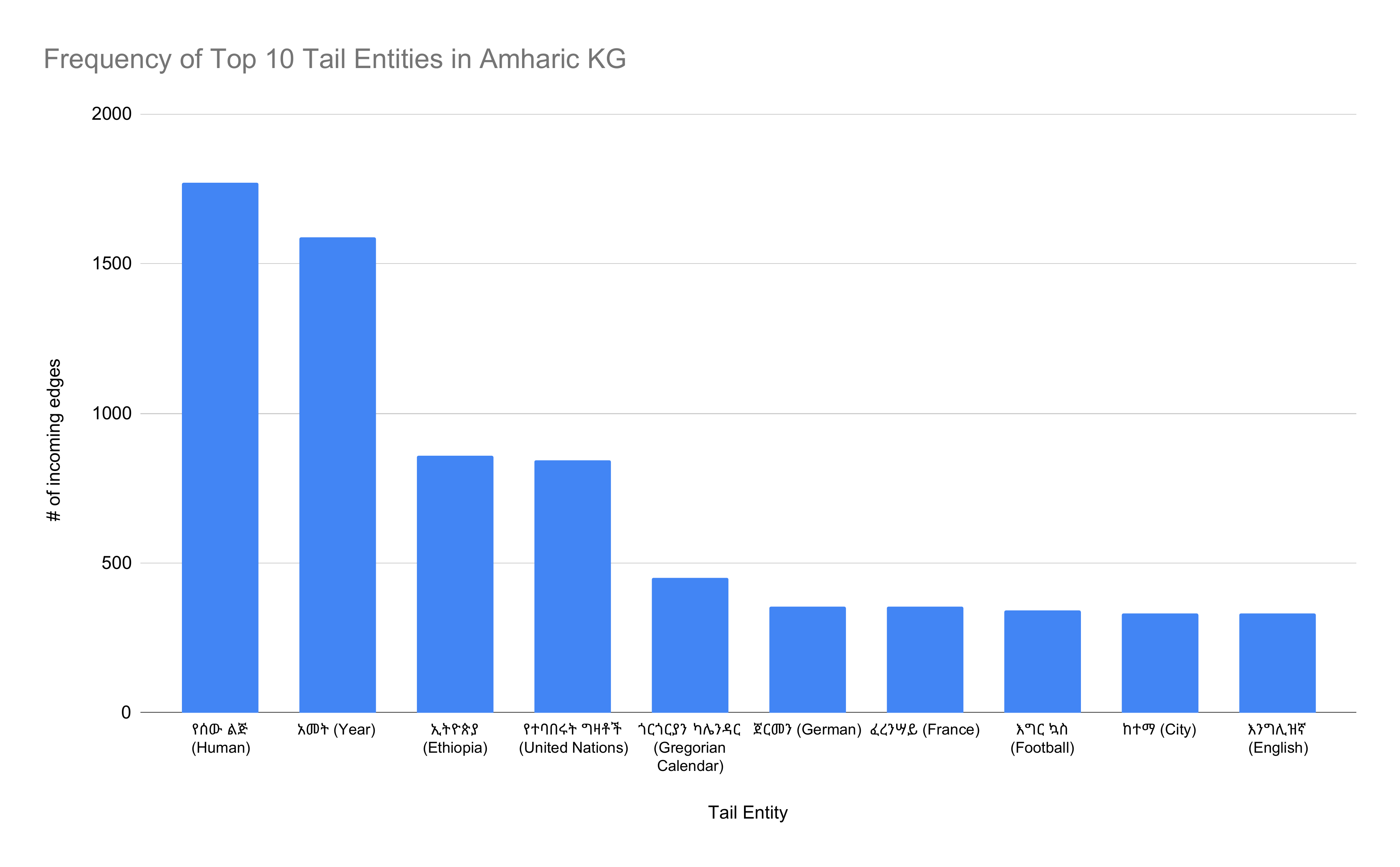}
        \label{fig:b}}
    \caption{Figure showing top 10 head and tail entities in Amharic KG. English translations for entities are provided in parentheses. }
    \label{fig:top_10_amh}
\end{figure*}

\begin{figure*}
    \centering
  \subfloat{%
        \includegraphics[width=0.5\textwidth]{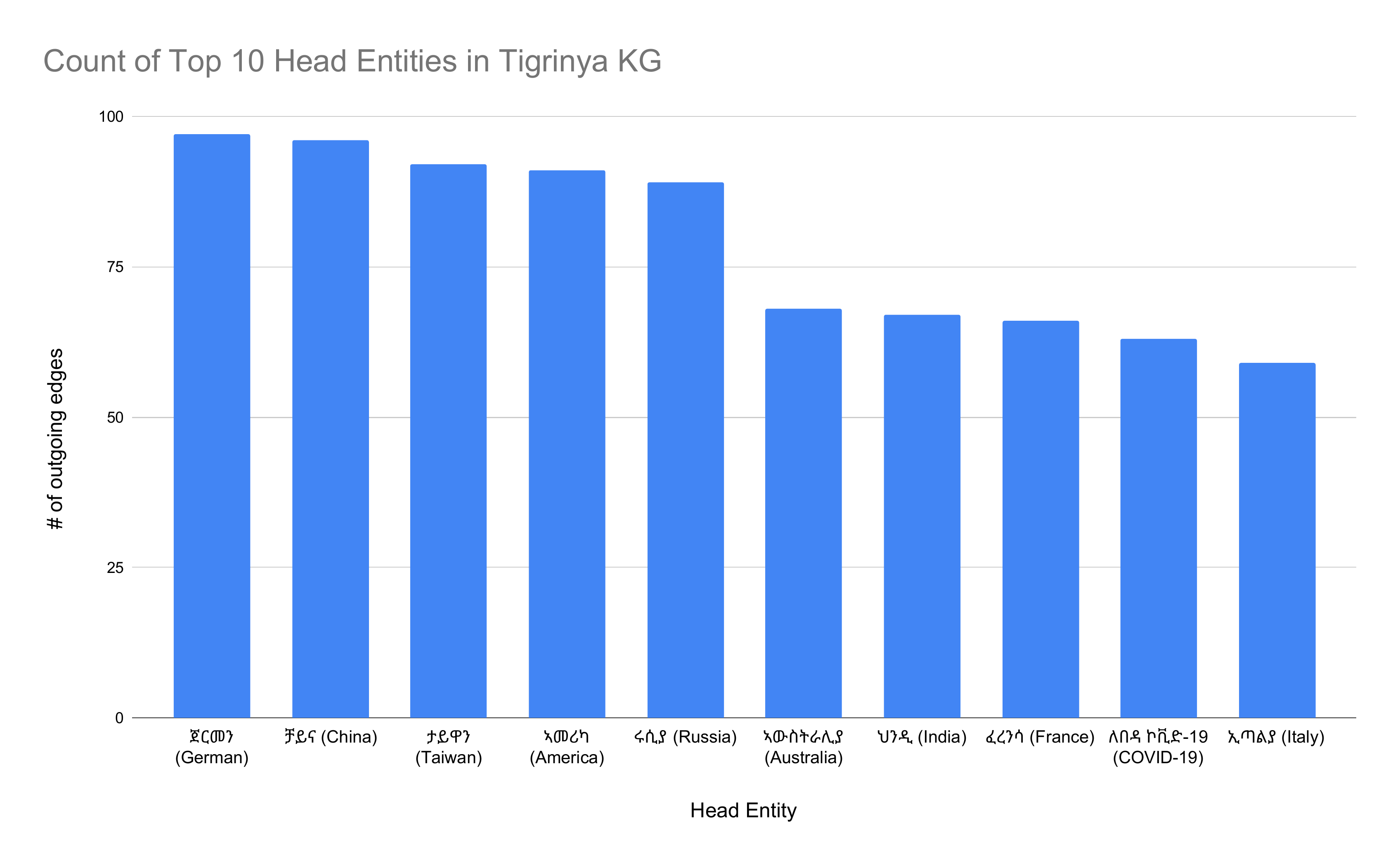}
        \label{fig:a}}
    \subfloat{%
        \includegraphics[width=0.5\textwidth]{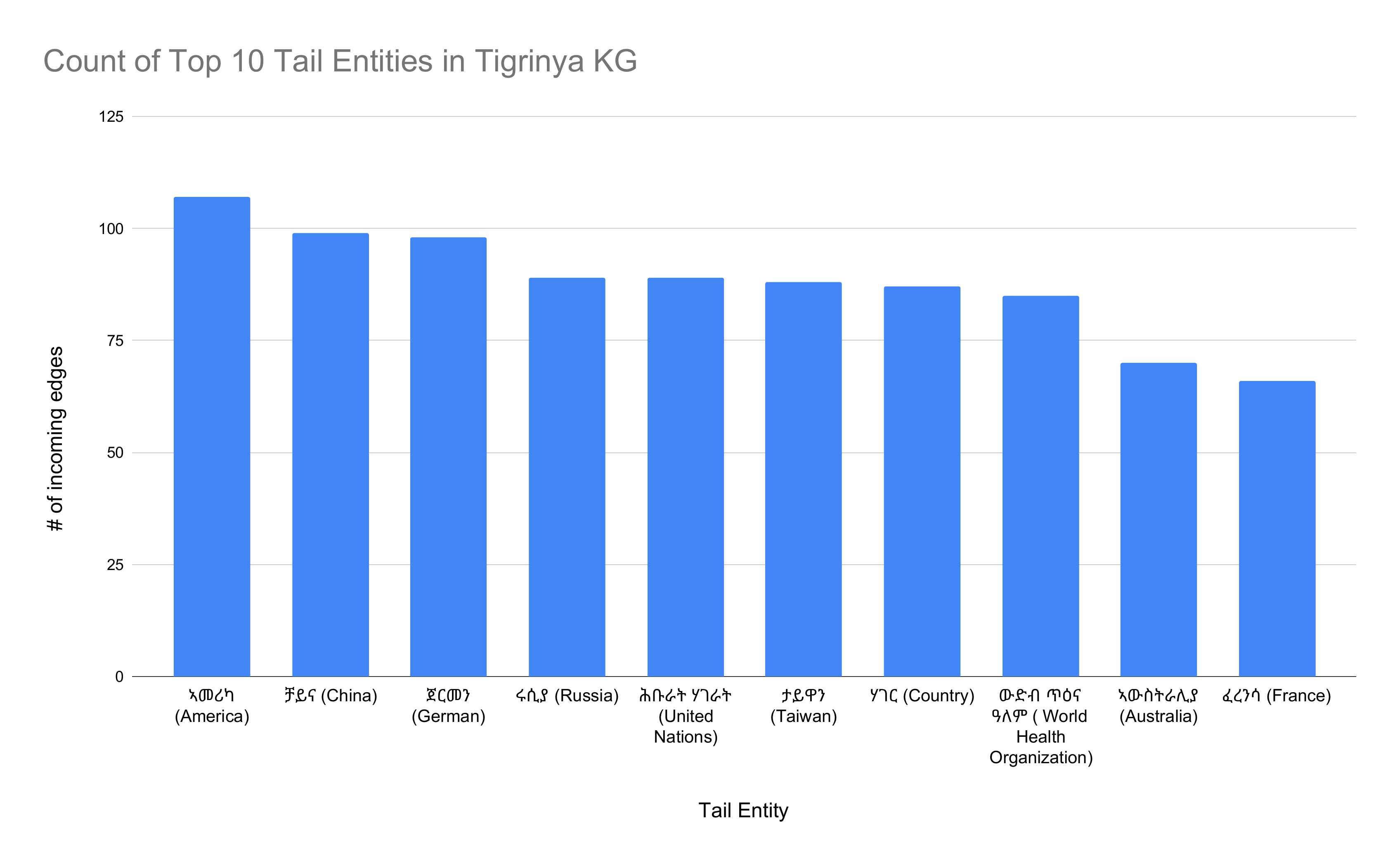}
        \label{fig:b}}
    \caption{Figure showing top 10 head and tail entities in Tigrinya KG. English translations for entities are provided in parentheses. }
    \label{fig:top_10_tir}
\end{figure*}

\begin{figure}
    \centering
    \includegraphics[width=\linewidth]{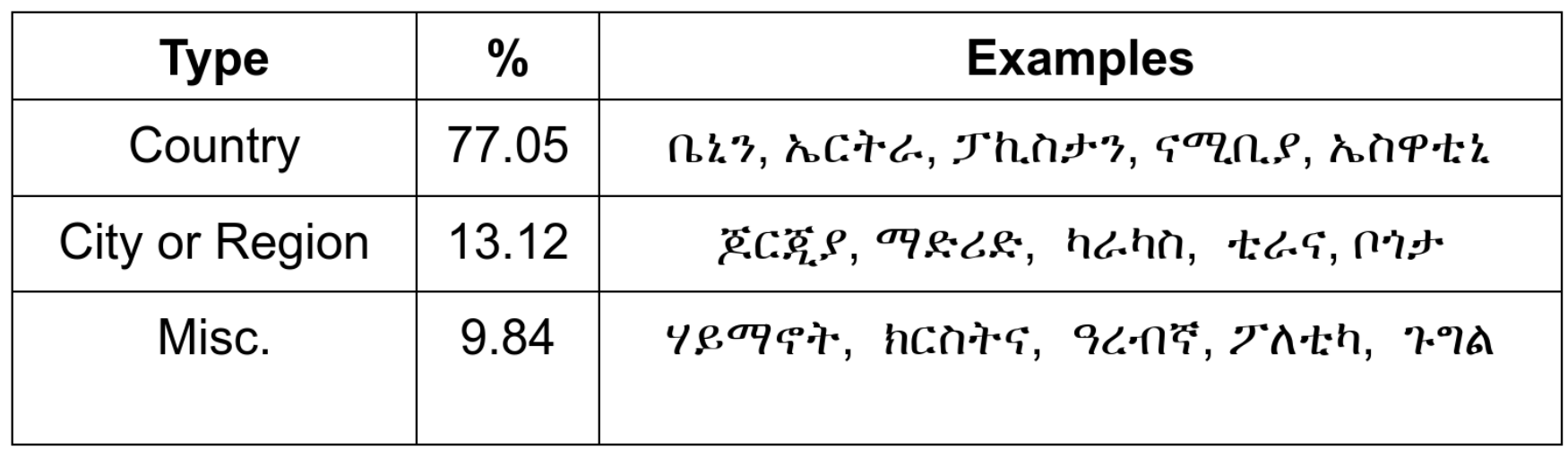}
    \caption{Tail entities that are spelled the same in Amharic and Tigrinya, allowing for shared representation in the model. We find that most of the shared tail entities are names of countries, cities and regions. }
    \label{fig:shared_entities}
\end{figure}

As detailed in Table \ref{tab:dataset}, there is a difference in the percentage of head and tail entities in the target languages that are covered by the transfer languages. Additionally, the number of Wikipedia articles available for each of the four languages of study varies significantly (see Table \ref{tab:dataset}. We took a deeper look at the entities that exist in the target language but do not exist in the transfer language. For the Amharic KG, of the 797 head entities that do not have textual representation in English, 268 have the tail entity ``Human'' and are names of individuals. We observe that 113 of the individuals are names of famous Ethiopians like Birtukan Dubale or Bahta Gebrehiwot or Ethiopian writers like Mimi Sebhatu and Sheh Tolha Jafar. Those entities are not covered by Tigrinya or Arabic. Of the 3125 head entities in Amharic KG not covered by Arabic, 592 have tail entity ``human'' and are names of individuals including famous Ethiopians as well as writers like Harold MacGrath and James Sallis which do exist in English KG. 
In the Tigrinya KG, of the 16 head entities that do not exist in Arabic KG, 7 of them are covered by Amharic KG and include traditional musical instruments of Ethiopia and Eritrea and locations in Ethiopia.

We also looked at the top 10 most frequent head and tail entities in the two KGs. As Figures \ref{fig:top_10_amh} and \ref{fig:top_10_tir} show, The top 10 head entities are mostly countries for both KGs. COVID-19 is also an entity that appears in the Top 10 for both KGs. In terms of top 10 tail entities, we see Amharic KG has ``human'' as the most frequent entity, indicating the KG mostly includes information about individuals followed by ``year'' indicating there are a lot of head entities that are years. The Tigrinya KG top 10 tail entities are mostly comprised of country names, indicating the KG is mostly represents information about relationships between different countries. 
\section{Zero-Shot Prompts} \label{apn:zero_shot}
In this section, we give an example of the prompt we used for the zero-shot experiment in \textsection \ref{sec:parametric}. For both Aya and GPT-4o, we used the same prompt. For a given target language, we take the reformulated question, Q, and design our prompt as follows:

\begin{mdframed}
   Please provide an answer for the following [LANGUAGE] question. Please keep your response to three words maximum and output the answer ONLY. Question: [Q] ?
   
   Answer:
\end{mdframed}

We attempt to constrain the model output to the tail entity only by instructing the model to output the answer only and asking it to keep the prediction at a maximum of 3 words. With manual inspection, we observed that the model might output additional tokens or words. Hence, we adjusted our evaluation function to count a prediction as correct if it contains the tail entity. 

\subsection{Details on KGT5 Setting}
We compared our approach with the scheme used in KGT5~\cite{kgt5-saxena2022sequence} and KGT5-context~\cite{kochsiek-etal-2023-friendly}. When comparing the Description and One-Hop connection-based schemes for providing context, we found that the performance did not improve or the two target languages. We hypothesized this could be due to the fact that the knowledge graphs are too small for the models to learn good enough representations on their own and the requirement for structured and labeled data constrained how useful \cite{kochsiek-etal-2023-friendly} approach would be for these low-resourced languages. Table \ref{tab:kgt5_reason} corroborates this hypothesis. 
\begin{table}[]
    \centering
    \begin{tabular}{cp{0.6cm}p{0.6cm}p{0.6cm}p{0.6cm}}
    \toprule
     & \multicolumn{2}{c}{Tigrinya} & \multicolumn{2}{c}{Amharic} \\
     \midrule
    &\% con. & \% tail & \% con. & \% tail \\
    \midrule
       KGT5-Description  & 49.77 & 1.78 & 6.3 & 0.71\\
       KGT5-One-Hop  &  48.83 & 0.89 & 25.77 & 1.65 \\
    \end{tabular}
    \caption{Percentage of context extracted from Wikidata Description and through One-Hop connections along with percentage of how many of the contexts have the tail entity. (Con. refers to context.)}
    \label{tab:kgt5_reason}
\end{table}


\end{document}